# D2S-FLOW: Automated Parameter Extraction from Datasheets for SPICE Model Generation Using Large Language Models

Hong Cai Chen, *Member, IEEE,* Yi Pin Xu, and Yang Zhang

***Abstract*—In electronic design, engineers often manually search through extensive documents to retrieve component parameters required for constructing SPICE models, a process that is both labor-intensive and time-consuming. To address this challenge, we present an automated framework called D2S-FLOW that leverages large language models (LLMs) to extract electrical parameters from datasheets and generate SPICE models with high precision and efficiency, significantly reducing the need for manual intervention. Unlike traditional RAG systems, D2S-FLOW employs a workflow to enhance precision in handling unstructured documents and inconsistent naming conventions through three innovative mechanisms: Attention-Guided Document Focusing (AGDF), Hierarchical Document-Enhanced Retrieval (HDER), and Heterogeneous Named Entity Normalization (HNEN). AGDF narrows retrieval to user-selected documents, HDER utilizes document structure for precise parameter localization, and HNEN standardizes terminology via semantic inference. Experimental results demonstrate that the framework achieves an Exact Match (EM) of 0.86, an F1 score of 0.92, and an Exact Correctness (EC) of 0.96, outperforming the strongest baseline by 19.4%, 5.7%, and 13.1%, respectively. Additionally, it reduces API token consumption by 38% and minimizes the irrelevant information ratio to 4%, showcasing substantial improvements in resource efficiency. This research provides an effective automated solution for circuit design.**

## I. Introduction

THE rapid advancement of LLMs has significantly propelled the field of natural language processing (NLP), particularly in tasks such as document retrieval and question answering [1], [2]. These models exhibit immense potential for enhancing electronic design automation EDA. LLMs have demonstrated diverse applications within EDA. For example, LLM4EDA [3] automates multiple steps in the EDA process by combining text and multimodal data such as circuit diagrams and code, significantly simplifying the design workflow. ChatEDA [4] further showcases how LLMs can generate processor design scripts and improve EDA efficiency by verifying accuracy through automated tools. Additionally, "The Dawn of AI-Native EDA" [5] discusses the application of AI-driven circuit models in circuit analysis and design, especially in standard cells and analog circuits, highlighting their advantages. Lastly, the AutoMage model [6] validates its high precision and reliability in EDA script generation tasks by comparing it with LLMs like GPT-4. DocEDA [7] integrates object detection, OCR, and LLM inference to automate the extraction of circuit diagrams and electrical parameters from documents, significantly enhancing the efficiency of circuit design document processing. Overall, these models primarily focus on design optimization and code generation in QA systems, promoting intelligent design development in EDA.

While, despite of design assistance, effective parameter extraction is also crucial for design processes and ensuring the accuracy of chip models [8], [9]. Engineers often consult extensive documentation to extract component parameters needed for constructing circuit simulation SPICE models. Manually searching through lengthy documents for parameters is time-consuming and labor-intensive [10], [11]. Automating parameter extraction and generating corresponding SPICE models would greatly enhance electronic design efficiency and increase EDA automation levels [12], [13]. LLMs possess a high capacity for document analysis and information extraction, offering potential to automate the parameter extraction process by accurately identifying and extracting model specifications and parameters from technical documents [14], [15].

Despite the advanced capabilities of LLMs, engineers face considerable challenges when extracting parameters from chip technical documentation. These documents are often lengthy and complex, making manual retrieval cumbersome. While methods like Retrieval-Augmented Generation (RAG) have shown promise in technical document analysis, directly applying LLMs does not efficiently facilitate parameter extraction due to the inherent complexity of chip documentation. This limitation underscores the necessity for innovative strategies to augment the parsing capabilities of LLMs in this specific context [6].

There are mainly two techniques to improve LLMs for

This work is funded by, China Postdoctoral Science Foundation (No. 2022M723913) and Fundamental Research Funds for the Central Universities (No. 3208002309A2). (Hong Cai Chen and Yi Pin Xu are co-first author. Corresponding author is Yang Zhang)

Hong Cai Chen and Yi Pin Xu are with School of Automation, Southeast University, Nanjing 210096, China. (email: 1455120321@qq.com, chenhc@seu.edu.cn).

Yang Zhang is with College of Advanced Interdisciplinary Studies, National University of Defense Technology, Changsha, China. (email: 16103271g@connect.polyu.hk)

specific applications: prompting and tuning. Tuning methods require large datasets and computational resources for training and are inflexible due to the need for continuous maintenance and updates. When introducing new data or features, or when network architectures change, retraining becomes necessary to adapt to new tasks or data variations. In contrast, prompting offers greater flexibility and cost-effectiveness but may struggle with complex tasks, necessitating meticulous design for optimal results [16].

The Chain-of-Thought (CoT) prompting method, which enhances LLMs' reasoning performance by generating incremental reasoning steps for complex tasks like mathematical problem-solving [17], may not suffice for parameter extraction from chip datasheets. This task involves navigating unstructured documents, inconsistent naming conventions, and high-dimensional data, where CoT alone often falls short. Auto-CoT [16] explored automatic generation of reasoning chains using similarity-based retrieval and zero-shot prompting, but it underperforms handcrafted chains on some datasets, indicating limitations in adaptability and precision for specialized tasks like parameter extraction.

To address these challenges, we propose a workflow-based framework called D2S-FLOW that decomposes the complex task of parameter extraction and model generation into manageable sub-tasks, leveraging multiple LLMs in a coordinated manner. This approach enhances the parsing ability of document parameters by distributing the workload across specialized models, each handling specific aspects of the extraction process. Our framework incorporates three innovative mechanisms: Attention-Guided Document Focusing (AGDF), Hierarchical Document-Enhanced Retrieval (HDER), and Heterogeneous Named Entity Normalization (HNEN). These mechanisms collectively improve the efficiency and accuracy of parameter extraction without requiring model fine-tuning, offering a robust and scalable solution.

The main contributions of this work are as follows:

- We introduce a novel workflow-based framework integrating AGDF, HDER, and HNEN to tackle the challenges of parameter extraction in EDA. AGDF reduces retrieval scope by focusing on user-selected documents, HDER leverages document structure for precise parameter localization, and HNEN standardizes inconsistent terminology through semantic inference.
- The proposed framework seamlessly integrates parameter extraction with SPICE model generation, automating the process from document analysis to model creation, significantly reducing manual effort and enhancing design efficiency. We performed simulation experiments to verify the functionality of the generated SPICE models, ensuring their practical applicability in circuit design.
- To validate the framework's effectiveness, we proposed a dataset and conducted comparative experiments, ablation studies, and model-agnostic performance evaluations, alongside proposing a new metric, Exact Correctness (EC), tailored to assess semantic correctness in chip parameter extraction scenarios.

The rest of the paper is organized as follows. In Section II, we review related works. Section III details our methodology. Section IV presents comprehensive experimental results and analysis. Finally, Section V concludes the paper.

## II. Related Works

Parameter extraction and modeling in EDA have been extensively studied, with various approaches proposed to address the challenges of efficiency and accuracy.

Traditionally, sparse vector techniques such as TF-IDF and BM25 have been widely used for document retrieval [17]. However, these methods often fall short in capturing semantic similarities due to their reliance on term frequency. Recent advancements have shifted towards dense text representations that allow for semantic-level modeling of textual similarity. For instance, Dense Passage Retrieval (DPR) [18] utilizes dual BERT models to generate embeddings for questions and text passages, significantly improving retrieval precision and enhancing the overall accuracy of end-to-end question-answering systems [19].

While these methods are effective in general contexts, they may not fully address the complexities inherent in technical document analysis within EDA, where specialized knowledge and precise parameter extraction are required.

Despite LLMs excelling in generating fluent and natural text, they are prone to hallucinations, generating content that is not faithful to the original source [20]. To mitigate this issue, RAG architectures have been introduced to combine parametric and non-parametric knowledge, aiming to improve generation accuracy and reduce hallucinations [21]. RAG integrates retrieval mechanisms to draw relevant content from external knowledge bases, thereby reducing the risk of inaccuracies. RAG has been extensively studied and applied in various specialized domains, as surveyed in [22], with notable effectiveness in medical [6, 23] and legal [24, 25] question answering. However, directly applying RAG to parameter extraction in EDA faces limitations due to the complexity and specificity of technical documents, and the necessity for high precision without extensive model fine-tuning [26].

Chain-of-Thought (CoT) prompting allows large language models (LLMs) to break down intricate tasks into sequential reasoning steps, improving outcomes in multi-step reasoning challenges [27]. This approach has demonstrated success in areas such as mathematical problem-solving and logical reasoning [28], and it holds particular significance for technical document analysis, where accurate parameter extraction is essential. However, despite its ability to bolster LLMs' reasoning for complex tasks, CoT may not be adequate for extracting parameters from chip datasheets [29].

The challenges, such as navigating unstructured documents, inconsistent naming conventions, and high-dimensional data, demand a more structured approach. Consequently, a workflow-based framework that decomposes the task into manageable sub-tasks and coordinates multiple LLMs becomes essential for ensuring accuracy and efficiency [30-33]. Our work addresses this gap by proposing a workflow-based framework integrating AGDF, HDER, and HNEN to enhance retrieval precision and multi-step reasoning for parameter extraction in EDA. This framework synergizes dense text representations and CoT reasoning, enhancing both retrieval precision and multi-step reasoning for parameter extraction.

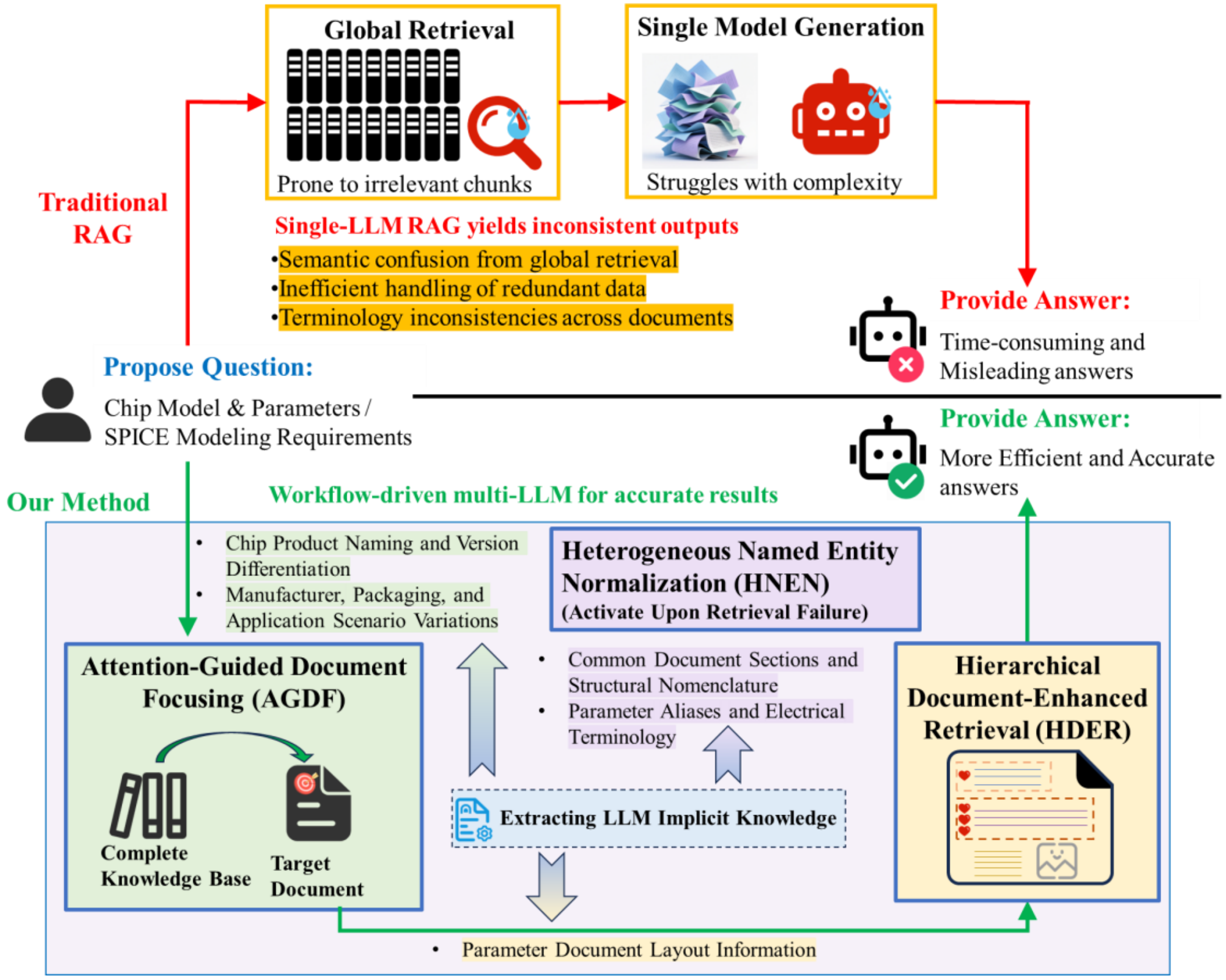

Fig. 1. The comparison of traditional RAG and the proposed DS-FLOW for chip parameter extraction. DS-FLOW decomposes the process into sub-tasks and integrated AGDF, HNEN and HDER strategies to obtain high efficiency.

## III. Methodology

This section introduces our proposed D2S-FLOW, a workflow-based framework that leverages multiple LLMs to automate the extraction of chip parameters from unstructured datasheets and generate SPICE models. Traditional RAG systems, typically reliant on a single LLM [34], struggle with limitations such as semantic confusion from global retrieval, inefficient handling of redundant data, and terminology inconsistencies across multi-vendor and multi-document contexts—issues that compromise the accuracy and efficiency of parameter extraction and SPICE model generation. To address these challenges, our framework employs a workflow-driven architecture that decomposes the task into distinct steps, each managed by specialized LLMs working collaboratively.

### *3.1 Framework Overview*

As shown in Fig. 1, this study presents D2S-FLOW, a workflow-based framework for chip parameter retrieval and SPICE model generation, designed to overcome the shortcomings of conventional RAG systems in multi-document, multi-vendor contexts. Anchored in workflow guidance and implicit knowledge extraction from LLMs, the framework integrates three core mechanisms: AGDF, HDER, and HNEN. These address semantic ambiguity from broad retrieval, limited fine-grained search precision, and cross-source naming disparities, enabling efficient and accurate parameter extraction and model synthesis.

The process begins with a user request for chip parameters or SPICE modeling, triggering an initial search across an extensive document set, followed by user selection of a target document. AGDF restricts subsequent retrieval to the chosen technical manual or data sheet, minimizing computational load and extraneous noise from wider searches. Then HDER employing LLM-derived knowledge via optimized prompts to identify and target parameter-rich sections within the document, extracting precise data and, where applicable, populating a SPICE model template. Following user validation of environmental conditions and parameter selections, the system delivers the extracted parameters or a finalized SPICE model, supporting circuit simulation.

Concurrently, HNEN ensures robustness against input errors and vendor-specific naming variations. By applying tailored prompts, the system standardizes equivalent terms and identifiers using LLM capabilities, preserving retrieval accuracy across diverse document sets. The combined effect of these mechanisms transforms unstructured chip data and user queries into high-precision structured outputs, achieving efficient parameter retrieval and SPICE model generation.

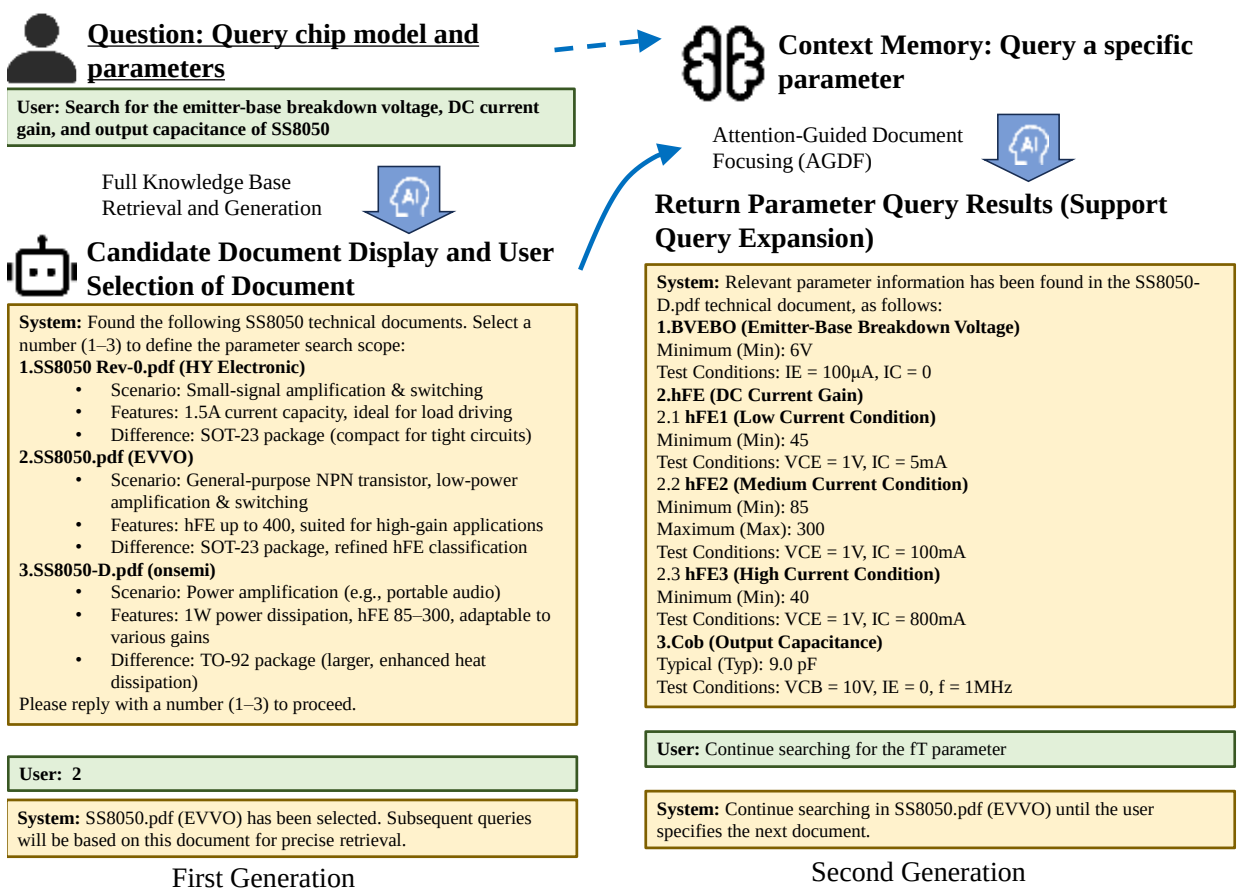

Fig. 2. Workflow of AGDF.

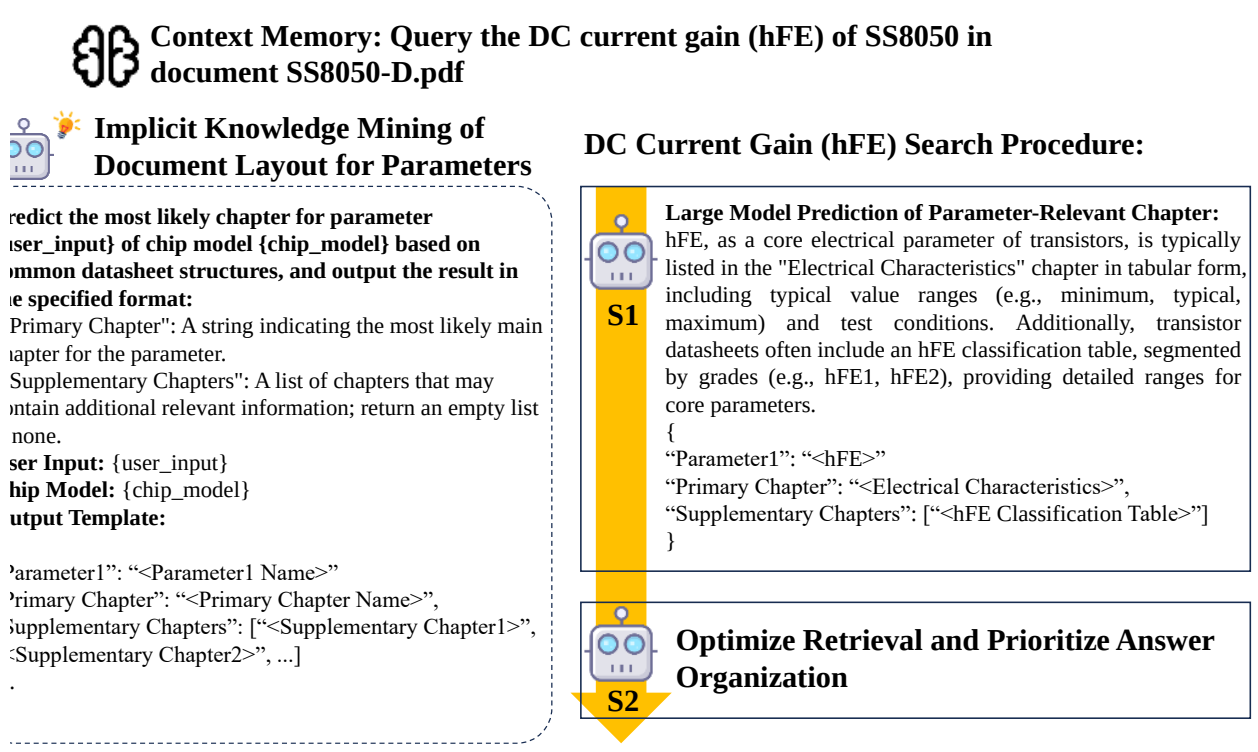

Fig. 3. Workflow of HDER.

## 3.2 *Parameter Extraction Methods*

Parameter extraction is the key link between user queries and structured outputs in the framework. This section details the three innovative mechanisms—AGDF, HDER, and HNEN. These mechanisms enhance retrieval precision and efficiency by leveraging user interaction, document structure, and semantic normalization.

### *A. Attention-Guided Document Focusing (AGDF)*

Traditional RAG systems perform global retrieval across expansive knowledge bases, often pulling in irrelevant documents that muddy the semantic waters and bog down computational efficiency. In contrast, the AGDF mechanism narrows this scope by dynamically limiting retrieval to a single datasheet chosen by the user, a process visually captured in Fig. 2. It starts when a user submits a parameter query—say, "Emitter-Base Breakdown Voltage, DC Current Gain, and Output Capacitance of SS8050." The system then fetches all relevant documents, assigning them labels like "1" for SS8050-A.pdf and "3" for SS8050-D.pdf, and asks the user to pick one—perhaps "3" for SS8050-D.pdf. From that point, all follow-up queries are tethered to this selected document, cutting out noise from unrelated sources. This sharpens retrieval precision, tackling a core weakness of traditional RAG systems where extraneous data often creeps in. The chosen document sticks for future queries unless the user switches chips, ensuring a consistent information stream. By weaving user interaction into the process, AGDF boosts accuracy and gives users reins over extraction, setting a sturdy stage for later mechanisms like HDER and HNEN. The perks are clear: it slashes irrelevant content by sticking to one document, lets users steer the system to their exact needs, and trims computational fat by tightening the retrieval focus.

Formally, while traditional RAG generates outputs as $y \sim p(y|C_x)$ with a broad context $C_x$ that often includes irrelevant information, AGDF refines this by focusing on a user-selected document, defining $C_x{}' = Focus(C_x)$ . Consequently, the output is generated as $y \sim p(y|C_x{}')$. This focused context significantly reduces noise, enhancing retrieval precision over conventional approaches.

### *B. Hierarchical Document-Enhanced Retrieval (HDER)*

Datasheets tend to scatter parameter details across various sections, and traditional full-text retrieval often drags in redundant or less critical data. The HDER mechanism steps in to streamline this, tapping the LLM's pre-trained smarts to pinpoint and rank relevant sections, as shown in Fig. 3. It works by borrowing from chain-of-thought prompting techniques [35], where the LLM uses short prompts to guess which sections hold the goods—for instance, pegging "Electrical Characteristics" as the go-to spot for "DC Current Gain (hFE)" and "hFE Classification Table" as backup. These guesses get tagged as labels, like {"Primary Section": "Electrical Characteristics," "Supplementary Section": "hFE Classification Table"}, and retrieval zeros in on just those areas, dodging irrelevant clutter. When generating answers, HDER splits the haul into primary info (the core numbers) and supplementary bits (extra context). Take "hFE" from the SS8050 datasheet: it pulls main values from "Electrical Characteristics" and adds classification details from "hFE Classification Table," delivering a tight, organized response. This layered method squeezes more from resources and lifts answer quality over clunky full-text approaches. It shines in precision by pre-sifting sections for only the good stuff, keeps outputs clear by sorting primary from supplementary data, and scales better by lightening the processing load with targeted retrieval.

HDER further improves the extraction process by incorporating implicit knowledge $E$, such as section priorities, and user input ( u ), building upon the focused context $C_x{}'$. This results in an enhanced output modeled as $y \sim p(y|C_x', E, u)$. By leveraging document structure and user preferences, HDER achieves more precise parameter localization compared to standard full-text retrieval methods.

### *C. Heterogeneous Named Entity Normalization (HNEN)*

Naming hiccups—thanks to vendor quirks or user slip-ups—often trip up retrieval accuracy in RAG systems [36-38]. The HNEN mechanism smooths this out by standardizing terms with the LLM's built-in know-how, sparked by simple prompts, as seen in Fig. 4. and Fig. 5. It shines in two key cases. First, for chip model mix-ups, if a user types "mega328" instead of "ATmega328" (Fig. 4.), HNEN kicks in after a retrieval miss, using LLM prompts to deduce the right model from linguistic cues—no bulky rulebooks needed. Second, for parameter and section terms like "Collector-Base Breakdown Voltage" or "Electrical Characteristics" (Fig. 5.), it scans the document to align varied labels, ensuring steady retrieval across sources.

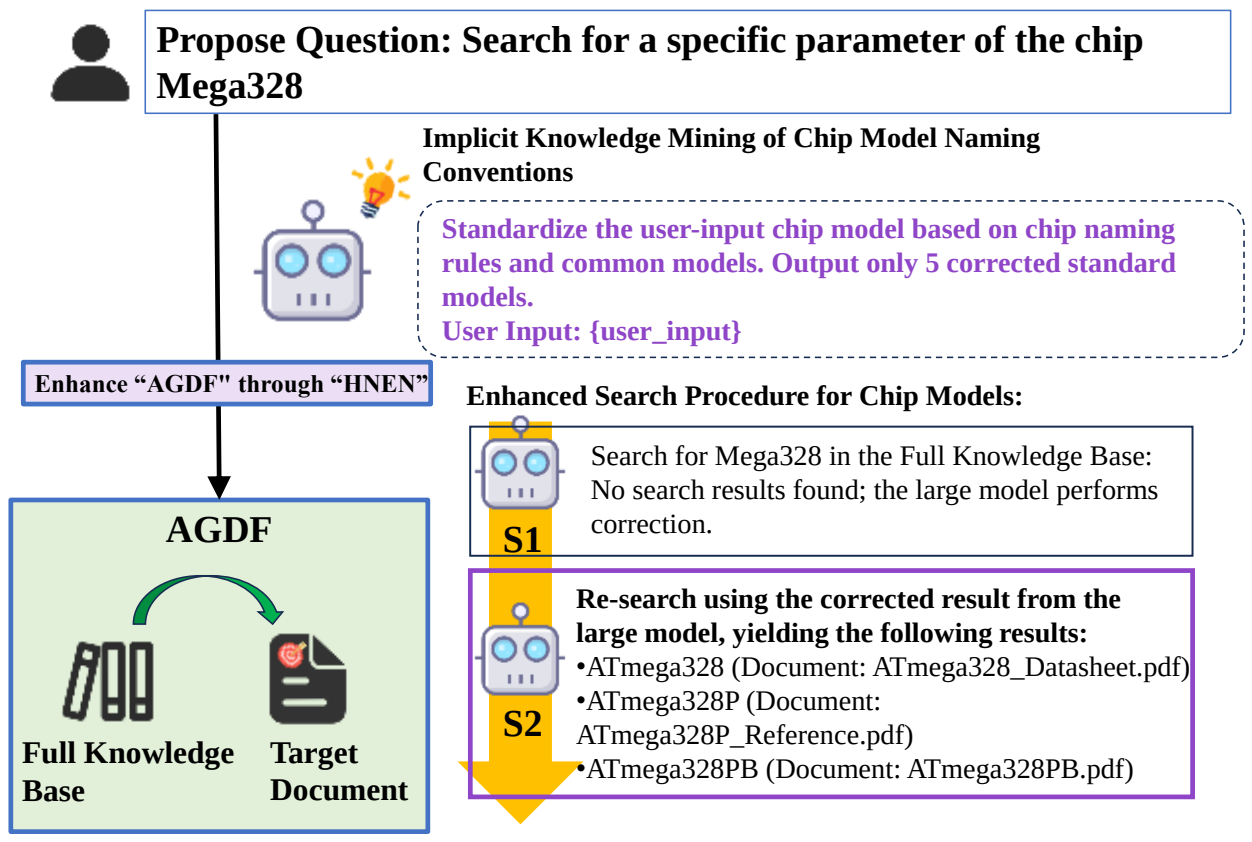

Fig. 4. Workflow for Enhancing AGDF through HNEN.

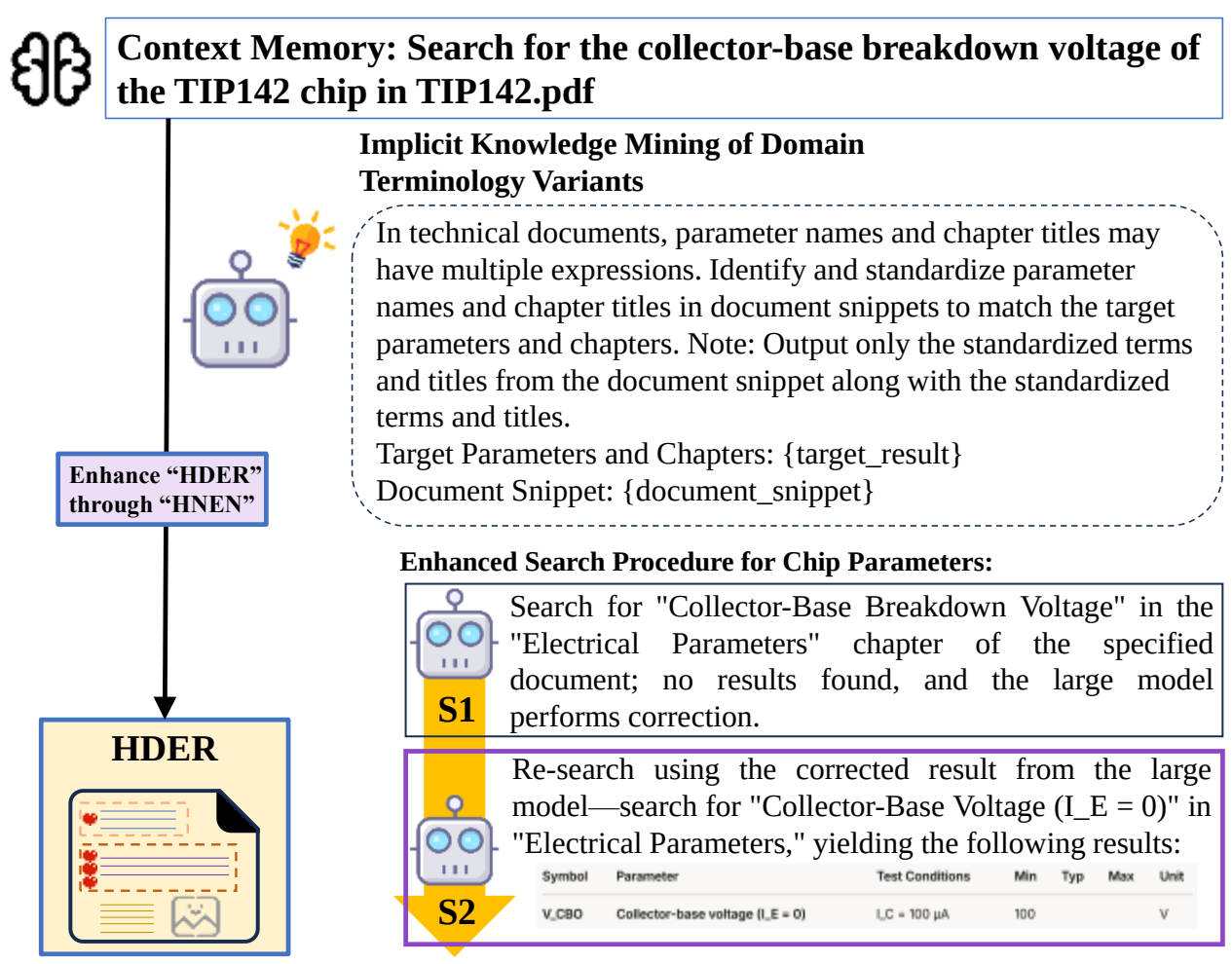

Fig. 5. Workflow for Enhancing HDER through HNEN.

This beefs up AGDF and HDER by making inputs tougher and adaptable to different datasheet flavors, all while dodging the upkeep headaches of rule-based fixes. HNEN stands out for its grit against input errors and naming twists, its knack for molding to document-specific lingo on the fly, and its lean approach, leaning on the LLM instead of sprawling rule sets.

To address terminological inconsistencies, HNEN normalizes the terms in the focused context, producing $C_x'' = Norm(C_x')$, where $Norm$ denotes the normalization operation applied to $C_x' = Focus(C_x)$. The output is then generated as $y \sim p(y|C_x'', E, u)$, ensuring robustness against variations in naming conventions across diverse documents and user queries.

### 3.3 SPICE Model Generation

Going beyond just pulling parameters, our framework automates SPICE model creation for circuit simulation, building on HDER with a template-driven twist, as laid out in Fig. 6. It kicks off with the LLM figuring out the chip type—think BJT or MOSFET—using its pre-trained brain, no rulebook required. Next, it picks a matching SPICE template and gathers all relevant parameters, including variants tied to different test conditions, then hands them to the user to choose from. Once the user locks in their picks, the system slots those values into the template, spitting out a full SPICE model. For AC simulations, it packs four small-signal templates covering diodes, BJTs, MOSFETs, and JFETs. Unlike traditional RAG setups that churn out stiff models, this method bends to user tweaks and parameter shifts, making it a practical gem for circuit design. It wins with customizability, letting users shape models to their specs; automation, flowing from inference to assembly with little hand-holding; and versatility, handling various device types and simulation needs.

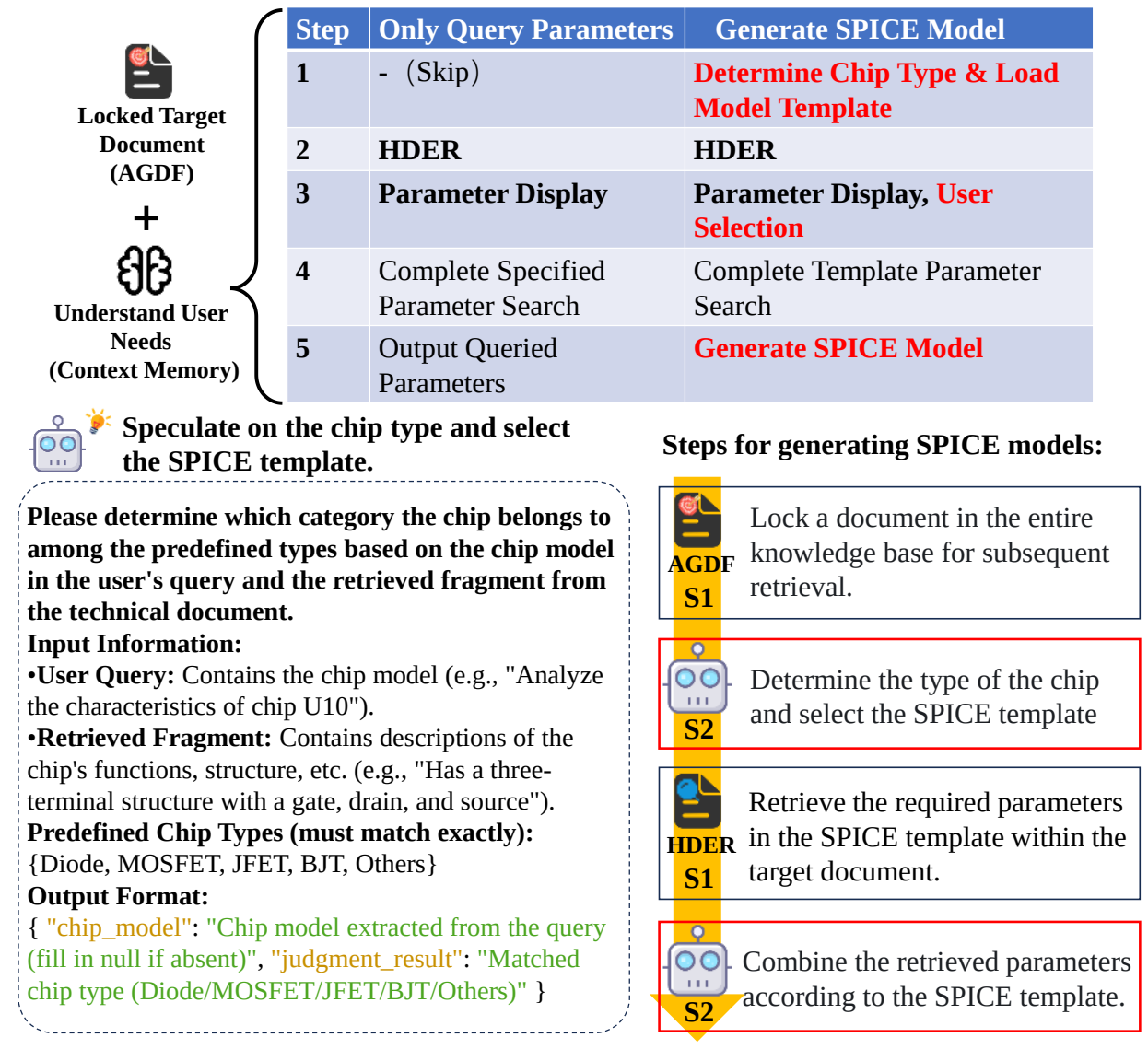

Fig. 6. Workflow for Enhancing AGDF through HNEN.

## IV. Experiments and Results

This section presents a thorough experimental evaluation of the proposed workflow-based framework for automated chip parameter extraction and SPICE model generation from unstructured datasheets. The experiments are designed to assess the framework's performance in terms of extraction quality, resource efficiency, robustness, and practical utility in generating functional SPICE models for circuit simulation. The evaluation encompasses dataset construction, query design, baseline comparisons, experimental setup (detailed experimental configuration in Appendix), parameter extraction results, ablation studies, cross-model performance analysis, and SPICE model validation. All reported results are averaged over three independent test sets to ensure statistical reliability.

### 4.1 Dataset and Evaluation Setup

To provide a robust foundation for evaluation, we constructed a specialized dataset of 80 chip datasheets sourced from DigiKey and Mouser, featuring a diverse range of semiconductor devices from 18 prominent brands, such as Texas Instruments, STMicroelectronics, etc. The dataset encompasses four primary device types—diodes, BJTs, MOSFETs, and JFETs—diversified into 28 subcategories. Datasheets were selected using a stratified sampling strategy: 60% were evenly distributed across the device types for balanced representation, while 40% were allocated based on a logarithmic scaling of market product counts to ensure practical relevance. Vendor bias was controlled by limiting each vendor's contribution to 30% per device type, resulting in a diverse and representative corpus (see TABLE I and **Appendix A** for details).

TABLE I
STATISTICAL DISTRIBUTION OF DATASET SAMPLES

| Device Type | DigiKey Samples Equal Dist. | DigiKey Samples Market Dist. | Mouser Samples Equal Dist. | Mouser Samples Market Dist. | Total Samples |
|---|---|---|---|---|---|
| Diode | 6 | 5 | 6 | 5 | 22 |
| BJT | 6 | 4 | 6 | 4 | 20 |
| MOSFET | 6 | 3 | 6 | 3 | 18 |
| JFET | 6 | 4 | 6 | 4 | 20 |
| Total | 24 | 16 | 24 | 16 | 80 |

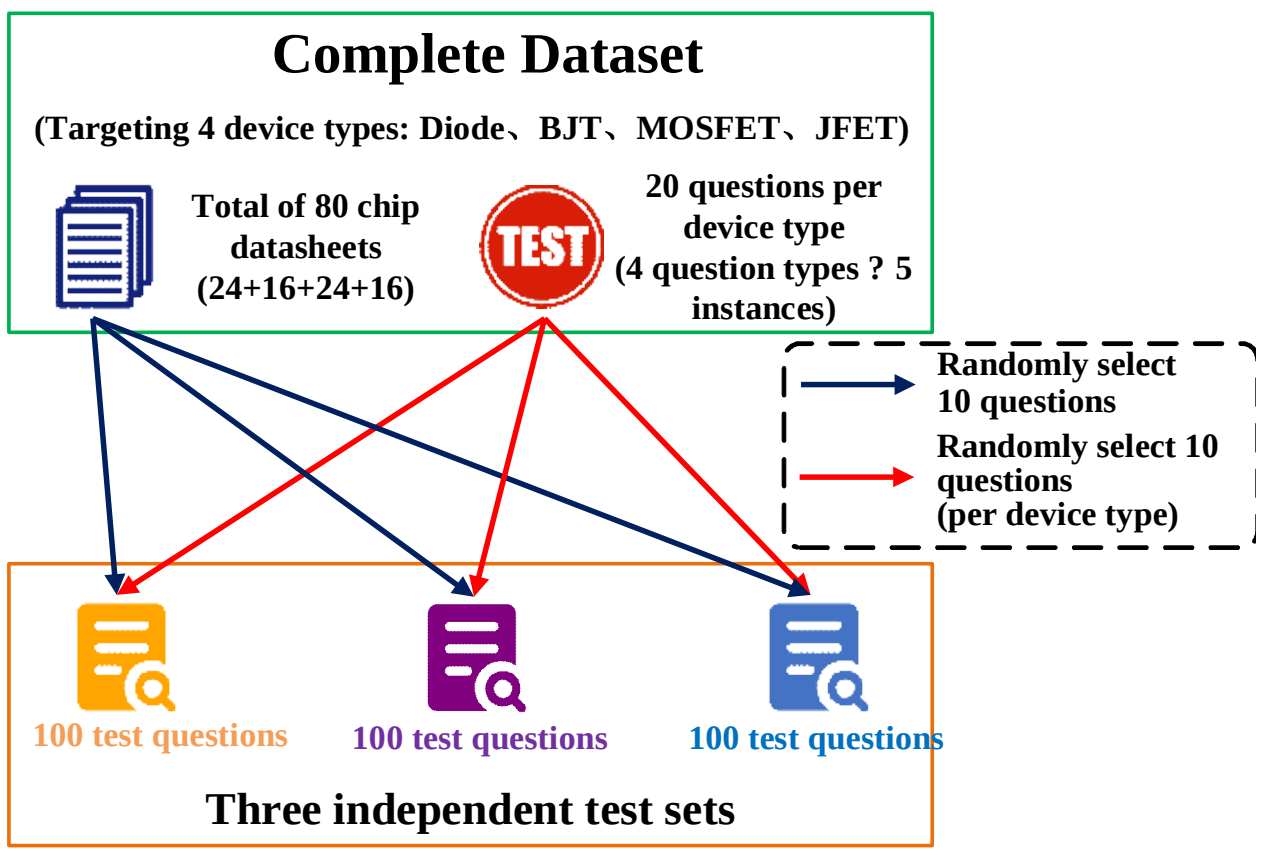

Fig. 7. Illustration of Test Set Construction Method

To test the framework's capabilities comprehensively, we designed 80 test queries categorized into four types:

- Basic parameter extraction, which involves queries for fixed parameters independent of conditions, such as "Extract the forward voltage of 1N4148";
- Complex condition-dependent extraction, which requires extracting parameters under varying conditions, for example, "Extract the threshold voltage of IRF540";
- Fuzzy input robustness, which tests the framework's tolerance to imprecise inputs, like "Extract SS850's emitter-base breakdown voltage," corrected to SS8050;
- Naming inconsistency handling, which evaluates the normalization of heterogeneous parameter names, such as "Extract BJT's V(BR)CBO," mapped to "V_CBO."

For each device type, five instances per query type were crafted, yielding 80 queries in total (see Fig. 7 and **Appendix B** for details). Three independent test sets were created, each comprising 10 randomly selected datasheets and 10 queries per device type (100 queries per set), balancing evaluation depth with computational feasibility.

Performance was evaluated across two dimensions: parameter extraction quality and resource consumption efficiency.

- Extraction quality: we used three metrics: exact matching (EM), which measures the proportion of queries with outputs identical to the ground truth; F1 Score, the harmonic mean of precision and recall, assessing partial correctness; and exact correctness (EC), a novel metric that employs LLMs to judge semantic equivalence, such as recognizing "2.5V at 25°C" as equivalent to "Vth=2.5V at 25°C," with further details provided in Table 3-4.

TABLE II
PARAMETER COMPARISON EXPERIMENT RESULTS

| Method | Parameter Search Quality Metrics: EM | F1 | EC | Resource Consumption & Performance Metrics: Response Time（s） | API Call Frequency | API Token Usage | Irrelevant Information Ratio |
|---|---|---|---|---|---|---|---|
| BM25+ Regex | 0.41 | 0.47 | 0.44 | **7** | - | - | 0.42 |
| RAG | 0.68 | 0.75 | 0.76 | 23 | 4.4 | 3845 | 0.24 |
| CoT | 0.72 | 0.87 | 0.84 | 29 | 5.2 | 4035 | 0.21 |
| D2S | **0.86** | **0.92** | **0.96** | 20 | **4.2** | **2493** | **0.04** |

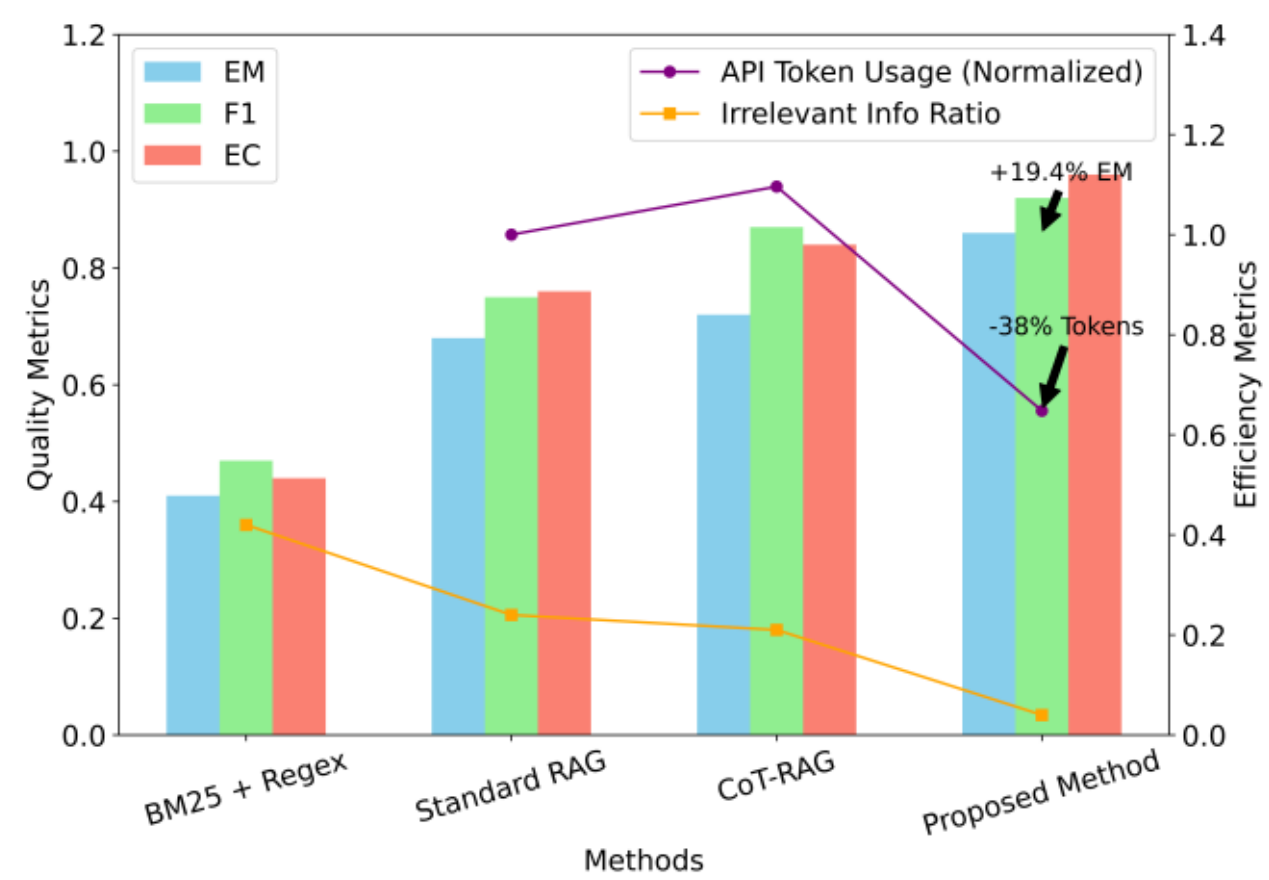

Fig. 8. Comparative Analysis of Parameter Extraction Methods

- Extraction Efficiency: it was quantified through Response Time in seconds, API Call Count, API Token Consumption, and the Irrelevant Information Ratio, which measures the fraction of extraneous output. These metrics collectively provide a comprehensive assessment of the framework's accuracy, robustness, and practicality.

Three baseline methods were implemented for comparison:

- BM25 + Regex, which combines BM25 retrieval with regular expression parsing, representing a traditional approach;
- Standard RAG, a basic Retrieval-Augmented Generation using LLMs without optimization;
- CoT-RAG, which enhances Standard RAG with Chain-of-Thought prompting for improved reasoning.

Detailed configurations of prompts and baseline methods can be found in **Appendix C & D**. Experiments were conducted on a high-performance workstation equipped with an Intel Xeon Platinum 8273CL processor, 128GB RAM, dual NVIDIA RTX 4090 GPUs, and a 2TB NVMe SSD, running Ubuntu 22.04 LTS, Python 3.10.12, Docker 24.0.0, RAGFlow v0.17.2, Elasticsearch v8.15.0, and PyTorch 2.3.0 (CUDA 12.6). Datasheets were converted from PDF to Markdown using the TextIn API (v2.1) to preserve tables. The framework utilized Qwen-3-8b for generation and Qwen-text-embedding-v2 for embeddings, employing a hybrid retrieval strategy with a keyword weight of 0.3, a similarity threshold of 0.2, and retrieval of the top-8 results.

### *4.2 Validation of Parameter Extraction and Analysis*

The proposed framework was evaluated against the baselines across all test sets, demonstrating superior performance in both quality and efficiency, as detailed in TABLE II and illustrated in Fig. 8. For parameter extraction quality, the framework achieved an EM of 0.86, an F1 Score of 0.92, and an EC of 0.95, significantly outperforming the baseline, CoT-RAG, which recorded 0.72, 0.87, and 0.84, respectively, with improvements of 19.4%, 5.7%, and 13.1%. The Standard RAG baseline yielded scores of 0.68, 0.75, and 0.76, while BM25 + Regex lagged at 0.41, 0.47, and 0.44, highlighting the framework's precision and semantic robustness, particularly evident through the EC metric. In terms of efficiency, the framework recorded a response time of 21 seconds, compared to 23 seconds for Standard RAG, 29 seconds for CoT-RAG, and 7 seconds for BM25 + Regex. Additionally, it achieved an API Call Count of 4.7 and API Token Consumption of 2493, which is 38% lower than Standard RAG's 3845. The Irrelevant Information Ratio was a mere 0.04, far below the 0.24 for Standard RAG, 0.21 for CoT-RAG, and 0.42 for BM25 + Regex, underscoring its ability to deliver focused and efficient outputs. These gains reflect the framework's precision in retrieving exact parameter values and its robust semantic understanding of parameter contexts within complex chip datasheets, where information is often dispersed across unstructured text, tables, or condition-dependent sections.

### *4.3 Validation of the Generated SPICE Model*

To validate the framework's ability to generate SPICE models, we extended it by integrating a generative module that identifies chip types, such as BJT or MOSFET, selects appropriate templates, extracts relevant parameters, and incorporates user-guided selection. For validation, 20 datasheets were randomly selected, and the generated SPICE models were tested in device-specific circuits: RC high-pass filters for diodes, common-emitter amplifiers for BJTs, and common-source amplifiers for MOSFETs and JFETs, with circuit configurations detailed in TABLE III. The models were evaluated based on functional performance criteria, such as achieving a cutoff frequency of approximately ($f_c \approx (2\pi RC)^{-1}$) for diodes and consistent gain for amplifiers, resulting in a perfect success rate of 20 out of 20, as shown in TABLE IV. Detailed simulation cases for each of the four device types are presented in **Appendix E**. In comparison, the Standard RAG baseline scored 12 out of 20, with five errors in template selection and three in parameter extraction, while CoT-RAG scored 13 out of 20, with four template errors and three extraction errors. These results underscore the framework's precision, attributable to accurate type identification and effective user interaction.

TABLE III
SPICE MODEL VALIDATION FOR FOUR DEVICE TYPES

| Device Type | SPICE Template | Parameter Source | Validation Circuit | Success Criterion |
|---|---|---|---|---|
| Diode | .model D1 D(IS CJO VJ M) | Datasheet | RC High-Pass Filter Circuit | $f_c \approx (2\pi RC)^{-1}$ |
| BJT | .model Q1 NPN (BF TF CJC CJE) | Datasheet | Common-Emitter Amplifier | Consistent low-frequency gain, minimal characteristic frequency error |
| MOSFET | .model M1 NMOS(VTO CGSO CGDO KP*) | Datasheet, *Empirical Value | Common-Source Amplifier | Consistent low-frequency gain, minimal 3 dB cutoff frequency error |
| JFET | .model J1 NJF(VTO CGS CGD BETA*) | Datasheet, *Empirical Value | Common-Source Amplifier | Consistent low-frequency gain, minimal 3 dB cutoff frequency error |

TABLE IV
SPICE ACCURACY VALIDATION SCORES FOR DIFFERENT METHODS

| Method | Score (out of 20) | Error Causes |
|---|---|---|
| Standard RAG | 12 | 5 template selection errors, 3 parameter search errors |
| CoT-RAG | 13 | 4 template selection errors, 3 parameter search errors |
| DS-FLOW | 20 | - |

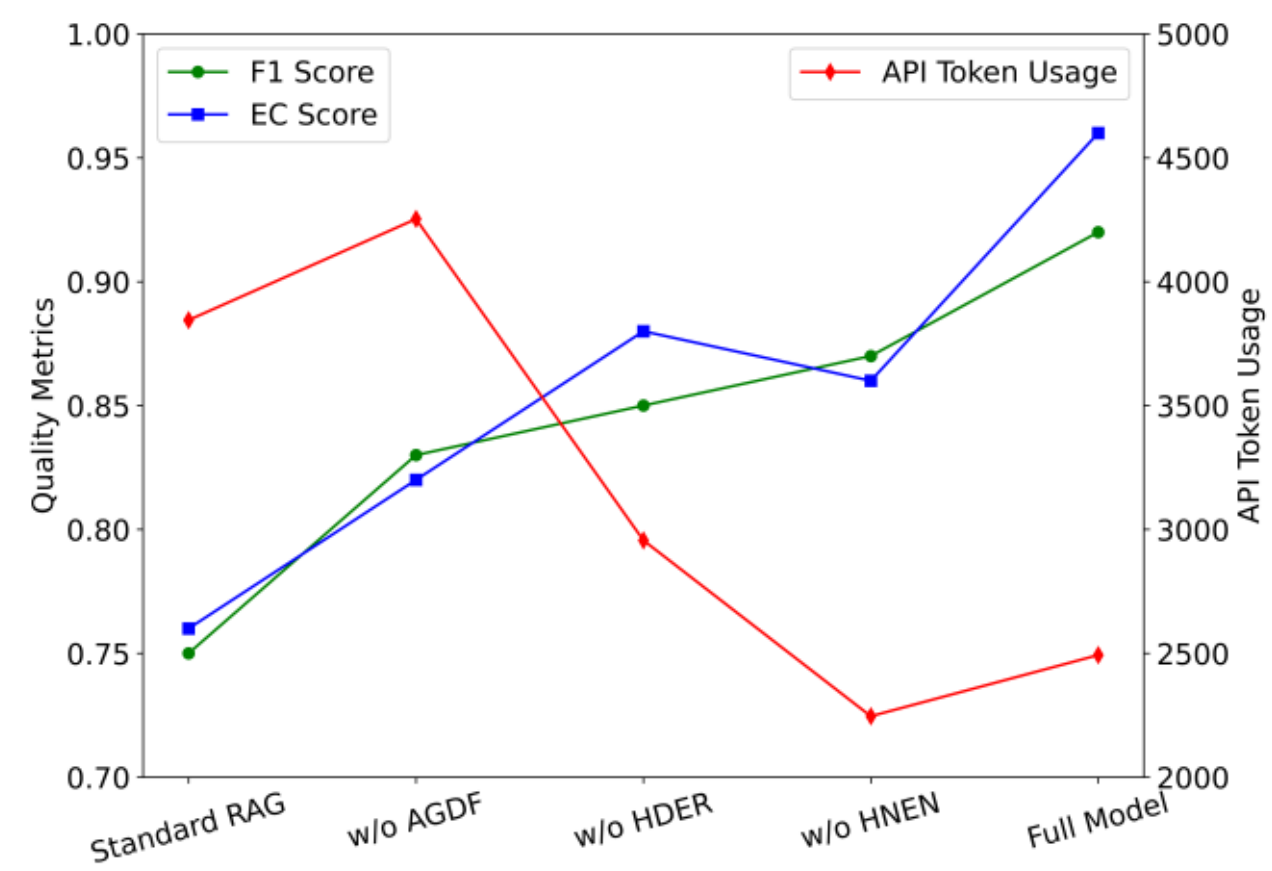

Fig. 9. Ablation Study of Mechanism Contributions

## V. DISCUSSION

The experimental results of this study highlight the efficacy of the proposed DS-FLOW for parameter extraction and SPICE model generation. This section examines these findings, elucidates the contributions of each mechanism, and discusses their implications for intelligent circuit design.

***Ablation Study***

An ablation study was conducted to quantify the contributions of the framework's three key mechanisms, as shown in Fig. 9. The workflow of the variants is detailed in **Appendix F**. The removal of AGDF results in a significant F1 score drop to 0.83 and an 81% surge in API token consumption to 4253, emphasizing its pivotal role in enhancing retrieval precision by confining the scope to a single, user-specified datasheet. This focus eliminates noise from irrelevant documents, a critical advantage in EDA where engineers prioritize specific components. HDER, by contrast, leverages the hierarchical structure of datasheets to prioritize parameter-rich sections, such as "Electrical Characteristics" or "Absolute Maximum Ratings." Its absence reduces F1 to 0.85 and EC to 0.88, underscoring its effectiveness in efficiently targeting key information and reducing extraneous retrieval. HNEN addresses naming inconsistencies across vendors, normalizing terms like "V(BR)CBO" to "V_CBO." While its removal yields a subtler decline (F1 to 0.87, EC to 0.86), its contribution enhances robustness in multi-vendor contexts. The full model, integrating all three mechanisms, achieves peak performance

TABLE V
PERFORMANCE COMPARISON RESULTS OF DIFFERENT LLM MODELS

| Model | Method | Parameter Search Quality Metrics | | Resource Consumption & Performance Metrics | | | |
|---|---|---|---|---|---|---|---|
| | | F1 | EC | Response Time(s) | API Call Frequency | API Token Usage | Irrelevant Information Ratio |
| Llama-3-8B | RAG | 0.65 | 0.67 | 30 | 5.2 | 4690 | 0.33 |
| | CoT | 0.73 | 0.75 | 35 | 5.7 | 4976 | 0.28 |
| | D2S | **0.9** | **0.93** | **21** | **4.3** | **2559** | **0.06** |
| Llama-3-70B | RAG | 0.75 | 0.77 | 65 | 1.2 | 6756 | 0.06 |
| | CoT | 0.83 | 0.85 | 78 | 1.2 | 7276 | 0.04 |
| | D2S | **0.95** | **0.97** | **45** | **0.7** | **4391** | **0.03** |
| Qwen 3-0.6B | RAG | 0.5 | 0.66 | 31 | 10 | 4052 | 0.36 |
| | CoT | 0.57 | 0.73 | 37 | 11.5 | 4251 | 0.32 |
| | D2S | **0.7** | **0.82** | **27** | **9** | **2430** | **0.11** |
| Qwen 3-1.7B | RAG | 0.58 | 0.69 | 33 | 9 | 4229 | 0.27 |
| | CoT | 0.66 | 0.76 | 39 | 10.5 | 4403 | 0.23 |
| | D2S | **0.79** | **0.87** | **25** | **8** | **2639** | **0.1** |
| Qwen 3-4B | RAG | 0.7 | 0.78 | 32 | 5.8 | 4439 | 0.22 |
| | CoT | 0.81 | 0.85 | 38 | 6.5 | 4738 | 0.18 |
| | D2S | **0.91** | **0.94** | **22** | **5** | **2638** | **0.06** |
| Qwen 3-8B | RAG | 0.75 | 0.76 | 23 | 4.4 | 3845 | 0.24 |
| | CoT | 0.87 | 0.84 | 29 | 5.2 | 4035 | 0.21 |
| | D2S | **0.92** | **0.96** | **20** | **4.2** | **2493** | **0.04** |
| Qwen 3-14B | RAG | 0.81 | 0.85 | 55 | 2.5 | 6627 | 0.12 |
| | CoT | 0.91 | 0.91 | 69 | 2.8 | 7202 | 0.1 |
| | D2S | **0.97** | **0.99** | **41** | **1.8** | **4003** | **0.02** |
| Qwen 3-32B | RAG | 0.83 | 0.87 | 72 | 1.9 | 7101 | 0.09 |
| | CoT | 0.91 | 0.93 | 86 | 2.1 | 7539 | 0.07 |
| | D2S | **0.98** | **0.99** | **52** | **1.3** | **4796** | **0.01** |

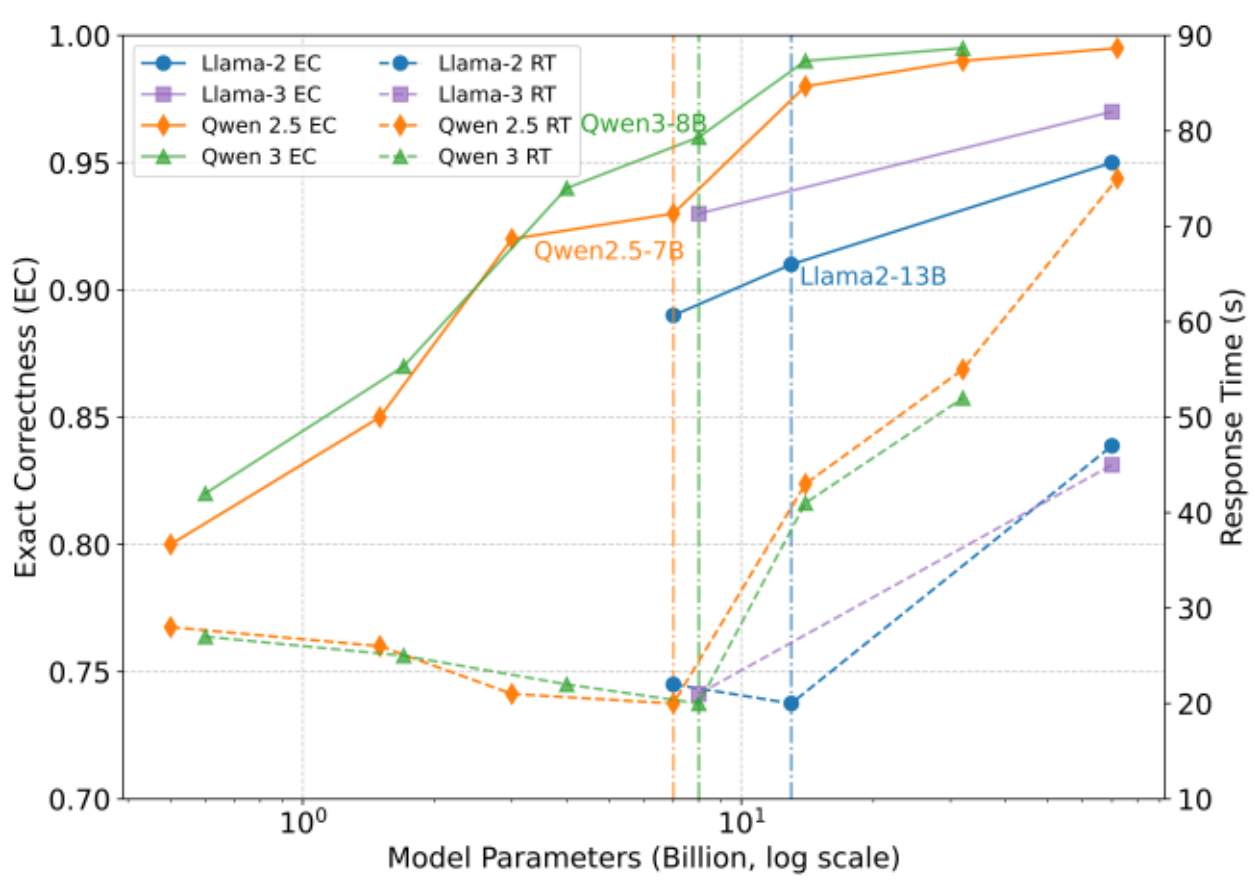

Fig. 10. Performance Scaling Laws: EC vs Model Size.

with an F1 of 0.94 and token usage of 2345, as annotated in Fig. 9.

Efficiency is another cornerstone of the framework's design, as evidenced in Fig. 9. It reduces API token consumption by 38% compared to Standard RAG and achieves an irrelevant information ratio of 0.04—substantially lower than baselines. These improvements stem from AGDF's targeted retrieval, HDER's structural prioritization, and HNEN's streamlined query processing. With an average response time of 21 seconds, the framework balances performance and practicality, making it suitable for real-time EDA applications and scalable integration into design workflows.

### *Generalizability Test*

The framework's generalizability was validated across the Llama-3 and Qwen 3 series, with detailed results presented in Table V, while additional models, including Llama-2 and earlier Qwen versions, are reported in **Appendix F**. Across all 18 tested models, D2S-FLOW consistently outperformed baseline methods. For instance, on the Llama-3-70B model, D2S-FLOW achieved an F1 score of 0.95 and an EC of 0.97, representing a 27% improvement in F1 over the Standard RAG baseline. Similarly, on the Qwen 3-32B model, D2S-FLOW attained an F1 score of 0.98 and an EC of 0.99, yielding an 18% gain over Standard RAG. Furthermore, D2S-FLOW significantly reduced resource consumption, with token usage decreasing by approximately 35% for Llama-3-70B and 32% for Qwen 3-32B, while irrelevant information ratios dropped to 0.03 and 0.01, respectively.

Cross-model analysis of the experimental data reveals three critical findings (Fig. 10): First, a performance-efficiency trade-off is evident, with mid-sized models like Qwen 3-8B achieving an optimal balance (EC=0.96, 20s), yielding a composite metric (EC / Response Time) of 0.048, outperforming smaller (Qwen 2.5-0.5B: EC=0.8, 28s) and larger models (Qwen 2.5-72B: EC=0.995, 75s). Second, generational advancements show newer models, such as Qwen 3-8B (EC=0.96, 20s) and Llama-3-8B (EC=0.93, 21s), surpassing predecessors like Qwen 2.5-7B (EC=0.93, 20s) and Llama-2-7B (EC=0.89, 22s), reflecting architectural enhancements. Third, efficiency peaks at 7B-14B parameters, with models like Qwen 2.5-7B and Llama-2-13B maintaining stable response times (~20s) before scaling penalties emerge. These results highlight Qwen 3-8B as the superior choice for high-demand tasks, with Qwen 2.5-7B and Llama-2-13B as efficient alternatives.

## VI. CONCLUSION

This research introduces an automated workflow-based framework called D2S-FLOW that harnesses LLMs to address the critical challenge of parameter extraction from chip datasheets and the generation of SPICE models for circuit design. By integrating three innovative mechanisms—AGDF, HDER, and HNEN—the framework overcomes limitations of traditional RAG systems, including semantic confusion from broad retrieval, inefficient processing of redundant data, and inconsistencies in terminology across diverse document sources.

The framework's effectiveness is evidenced by rigorous experimental evaluations, achieving an EM score of 0.86, an F1 score of 0.92, and an EC score of 0.96—outperforming the strongest baseline by 19.4%, 5.7%, and 13.1%, respectively. Additionally, it reduces API token consumption by 38% and minimizes the irrelevant information ratio to 4%, demonstrating superior resource efficiency. Validation experiments further confirm the practical utility of the generated SPICE models, with a 100% success rate in functional simulations across various device types, reinforcing the framework's value in EDA.

## APPENDIX A: DATASET CONSTRUCTION AND STATISTICS FOR TECHNICAL MANUALS

This appendix details the construction process, sampling strategy, and question design for the dataset used in this study. The dataset comprises 80 datasheets covering four electronic device types: diodes, bipolar junction transistors (BJTs), metal-oxide-semiconductor field-effect transistors (MOSFETs), and junction field-effect transistors (JFETs). Sourced from DigiKey and Mouser (40 each), these documents ensure diversity and authority.

A stratified sampling method was employed to balance device type representation and market distribution authenticity. Sampling consists of two parts: uniform allocation (60%, 48 datasheets) and market allocation (40%, 32 datasheets). Uniform allocation assigns 12 datasheets per device type (6 from DigiKey, 6 from Mouser), ensuring equal representation. Market allocation is based on logarithmic transformations of product counts from DigiKey and Mouser, reflecting real-world application patterns—e.g., diodes receive more samples due to higher product numbers. To enhance diversity and reduce bias, no single manufacturer exceeds 30% of a device type's samples (e.g., max 6 of 20 MOSFET datasheets).

The sample distribution is as follows:

- Diodes: 22 datasheets (DigiKey: 11 [6 uniform, 5 market, 272,665 products]; Mouser: 11 [6 uniform, 5 market, 268,384 products]).
- BJTs: 20 datasheets (DigiKey: 10 [6 uniform, 4 market, 31,069 products]; Mouser: 10 [6 uniform, 4 market, 13,009 products]).
- MOSFETs: 20 datasheets (DigiKey: 10 [6 uniform, 4 market, 64,190 products]; Mouser: 10 [6 uniform, 4 market, 22,269 products]).
- JFETs: 18 datasheets (DigiKey: 9 [6 uniform, 3 market, 3,212 products]; Mouser: 9 [6 uniform, 3 market, 982 products]).

This dual allocation ensures balance and market relevance, providing a robust foundation for parameter search system evaluation.

## APPENDIX B: DESIGN OF PARAMETER EXTRACTION TEST QUESTIONS

This study designed 80 test questions to assess the parameter search system, spanning four device types with 5 questions per type across four categories. Questions follow the format: "Query the {query parameter} of {chip model}," with chip models derived from specific datasheets. TABLE VI lists basic and complex condition parameters for basic and complex extraction tasks, while TABLE VII details uncommon parameter names for robustness testing, ensuring comprehensive and logical question design.

Basic extraction tests fixed parameter retrieval, reflecting core system performance. Complex extraction targets condition-dependent parameters (e.g., under varying temperatures), requiring precise identification.

TABLE VI
BASIC AND COMPLEX CONDITION PARAMETERS

| Chip Type | Basic Parameter Extraction | Complex Condition Parameter Extraction |
|---|---|---|
| Diode | Package Type | Breakdown Voltage |
| | Maximum Repetitive Reverse Voltage | Forward Voltage |
| | Power Dissipation | Reverse Leakage Current |
| | DC Forward Current and Repetitive Peak Forward Current | Total Capacitance |
| | Storage and Operating Junction Temperature Range | Reverse Recovery Time |
| BJT | Package Type | Breakdown Voltage |
| | Junction Temperature | Collector Cutoff Current |
| | Storage Temperature | DC Current Gain |
| | Collector Current | Switching Time |
| | Collector-Base Voltage | Collector-Emitter and Base-Emitter Saturation Voltage |
| MOSFET | Package Type | Drain-Source Breakdown Voltage |
| | Drain-Source and Gate-Source Voltage | Drain Leakage Current |
| | Pin Information | Junction-to-Ambient Thermal Resistance |
| | Source Current | On-Resistance |
| | Storage Temperature | Input Capacitance |
| JFET | Package Type | Gate Reverse Current |
| | Operating and Storage Temperature Range | Gate-Source Cutoff Voltage |
| | Drain-Source Voltage | Off-State Drain Current |
| | Drain-Gate Voltage | Zero-Gate Voltage Drain Current |
| | Thermal Resistance | Static Drain-Source On-Resistance |

## APPENDIX C: DETAILED CONFIGURATION OF PROMPT ENGINEERING

This appendix describes four systems designed for chip datasheet analysis and standardization using advanced prompt engineering.

- **Datasheet Analysis Assistant:** Processes retrieval results for chip models (e.g., "2n2222") and parameters (e.g., "current"), generating a numbered list of differences (e.g., "1. 2n2222a.pdf: 600mA, TO-92; 2. 2n2222-datasheet.pdf: 800mA, TO-18") for user selection.
- **Parameter-Section Mapping System:** Predicts datasheet sections for parameters (e.g., "voltage") in JSON format: {"Parameter": "Voltage", "Main section": "Electrical Characteristics", "Supplementary sections": ["Absolute Maximum Ratings"]}
- **Chip Model Standardization System:** Corrects and standardizes chip model inputs (e.g., "2n222" to "2N2222", "2N2222A") for compatibility with official documentation.
- **Parameter and Section Inference System:** Standardizes parameters and section titles from

TABLE VII
UNCOMMON PARAMETER NAMES AND STANDARDS

| Chip Type | Uncommon Name | Standard Parameter | Standard Parameter |
|---|---|---|---|
| Diode | Reverse Avalanche Voltage | Reverse Breakdown Voltage | Reverse Breakdown Voltage |
| | Peak Inverse Voltage (PIV) | Maximum Repetitive Reverse Voltage | Maximum Repetitive Reverse Voltage |
| | Thermal Power Dissipation Capacity (TPDC) | Power Dissipation | Power Dissipation |
| | Direct Current Forward Bias and Repetitive Peak Forward Surge Current (DCFBC and RPFS) | DC Forward Current and Repetitive Peak Forward Current | DC Forward and Repetitive Peak Current |
| | SATR and JOTR | Storage Temperature Range and Operating Junction Temperature Range | Storage and Junction Temperature Range |
| BJT | Non-Operating Temperature | Storage Temperature | Storage Temperature |
| | Collector Output Current | Collector Current | Collector Current |
| | CBB | Collector-Base Voltage | Collector-Base Voltage |
| | Off-State Current | Collector Cutoff Current | Collector Cutoff Current |
| | Ambient Temperature | Operating Temperature | Operating Temperature |
| MOSFET | DSB | Drain-Source Voltage | Drain-Source Voltage |
| | Drain Off-State Current | Drain Leakage Current | Drain Leakage Current |
| | Thermal Resistance from Junction to Air | Junction-to-Ambient Thermal Resistance | Junction-to-Ambient Thermal Resistance |
| | Source Electrode Current | Source Current | Source Current |
| | RDS(on) | On-Resistance | On-Resistance |
| JFET | Temperature Range for Operation and Storage | Operating and Storage Temperature Range | Operating and Storage Temperature Range |
| | Pinch-Off Voltage | Gate-Source Cutoff Voltage | Gate-Source Cutoff Voltage |
| | Drain Current at Cutoff | Off-State Drain Current | Off-State Drain Current |
| | Drain-Gate Bias | Drain-Gate Voltage | Drain-Gate Voltage |
| | RDS(static) | Static Drain-Source On-Resistance | Static Drain-Source On-Resistance |

document snippets, ensuring consistency across datasheet formats.

These systems collectively enhance datasheet processing for scientific research.

## APPENDIX D: DESIGN AND IMPLEMENTATION OF BASELINE METHODS

Three baseline methods were designed to evaluate the proposed approach, representing traditional information retrieval and RAG techniques. All use chip datasheets (PDF-to-Markdown) and identical environments for comparability.

- **Baseline 1**: Combines exact matching and structured parsing.
  - **Design**: Two-step retrieval—locate documents, then extract parameters with tables.
  - **Implementation**: Fuzzy matching (Levenshtein distance ≤ 2), BM25 (k1=1.5, b=0.75), regex (e.g., r'(\d+.\d+)\sV'), and a 50-variant synonym table (Table D).
  - **Pros**: Preserves context.
  - **Cons**: Limited by naming heterogeneity.
- **Baseline 2**: Standard RAG architecture.
  - **Design**: Semantic retrieval and LLM generation.
  - **Implementation**: 512-character chunks, Qwen embeddings, hybrid query (threshold 0.2, weight 0.3), Qwen-3-8B (temperature=0.7, max_tokens=50).
  - **Pros**: Handles naming variance.
  - **Cons**: May lose precision.
- **Baseline 3**: RAG with prompt engineering.
  - **Design**: Adds logical prompts to Baseline 2.
  - **Implementation**: Prompt with 3 steps (identify, locate, extract), Qwen-3-8B (temperature=0.5, max_tokens=100).
  - **Pros**: Enhanced accuracy.
  - **Cons**: Increased complexity.

## APPENDIX E: VALIDATION OF SPICE MODELS THROUGH SIMULATION EXPERIMENTS

This appendix provides detailed simulation experiments to validate the SPICE models for four semiconductor devices: diodes, bipolar junction transistors (BJTs), metal-oxide-semiconductor field-effect transistors (MOSFETs), and junction field-effect transistors (JFETs). These experiments, supporting Section 4.3, demonstrate the consistency between simulation results and theoretical expectations, confirming the correctness and usability of the SPICE models.

**E.1 Validation of Diode SPICE Model Parameters Using an RC High-Pass Filter**

**Objective:** This experiment verifies the zero-bias junction capacitance CJO = 4×10−12 F in the SPICE model of small-signal diodes (1N91x, 1N4x48, FDLL914, FDLL4x48 series) from ON Semiconductor, defined as:

.model D1N916 D(IS=25e-9 CJO=4e-12 VJ=0.7 M=0.5)

**Experimental Design:** An RC high-pass filter circuit, comprising a diode in series with a R = 1kΩ resistor, evaluates the junction capacitance under reverse bias. The theoretical cutoff frequency is:

$$f_c = (2\pi R C_j)^{-1}$$

**Experimental Procedure:**

1) Calculate Theoretical Junction Capacitance: At VR = 10V,

$$C_j = \frac{C_{JO}}{\left(1+\frac{V_R}{VJ}\right)^M} = \frac{4\times10^{-12}}{\left(1+\frac{10}{0.7}\right)^{0.5}} = 1.02\times10^{-12}F$$

2) Determine Theoretical Cutoff Frequency:

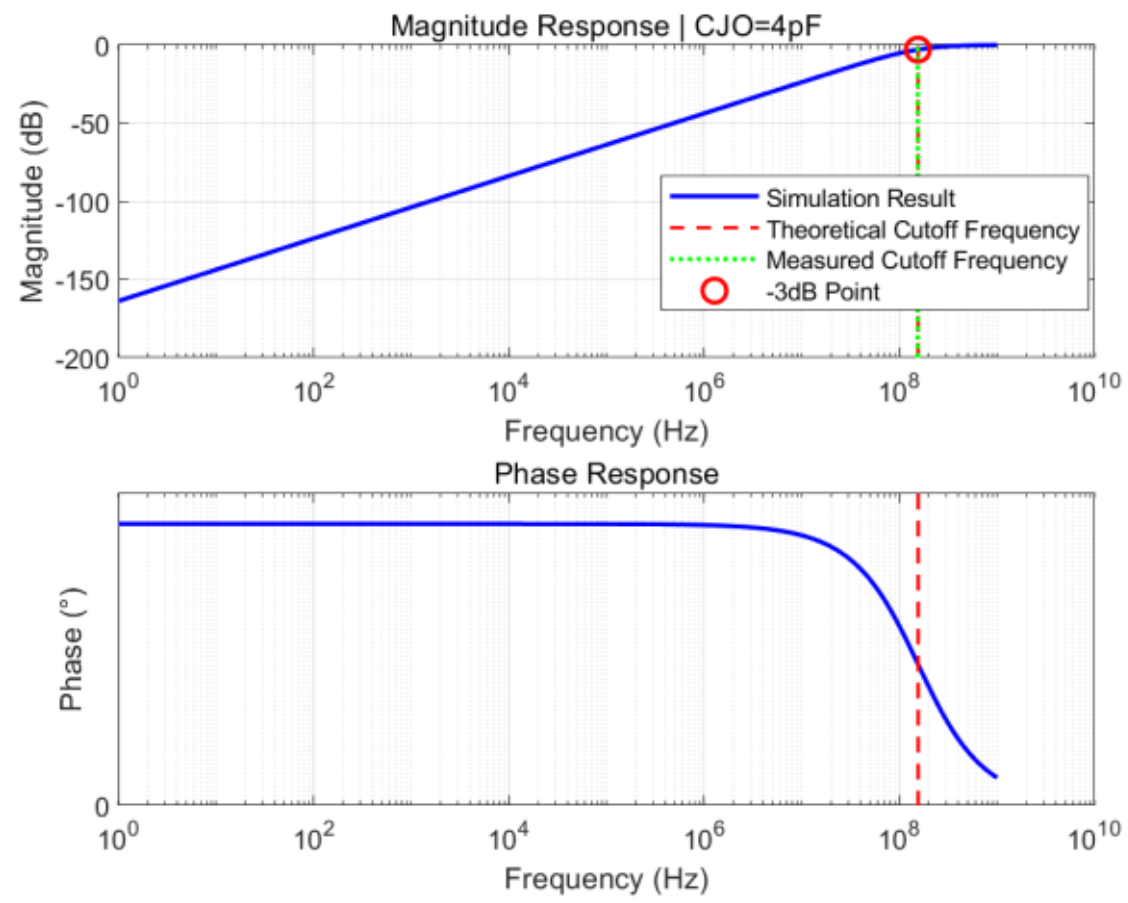

Fig. 11. Frequency response of the RC high-pass filter for diode validation.

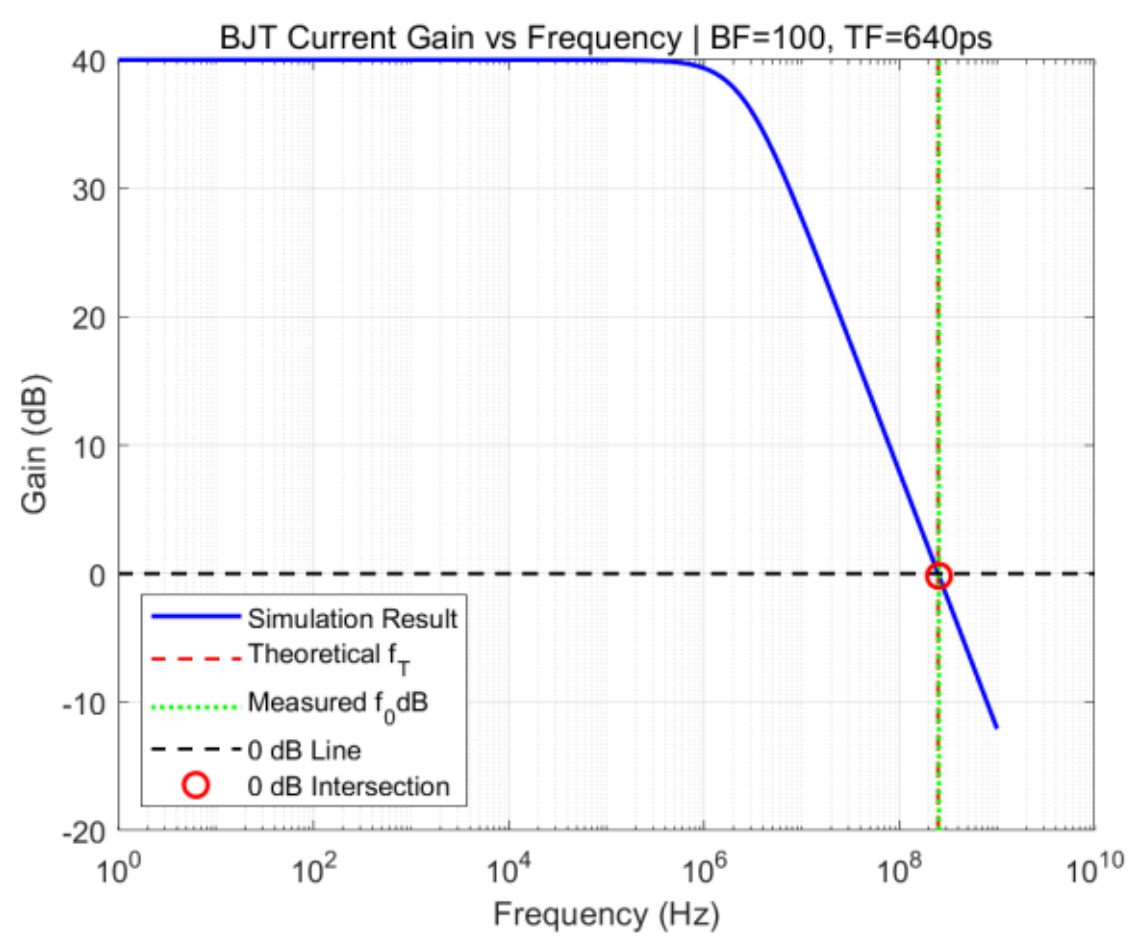

Fig. 12. Frequency response analysis of the BJT model

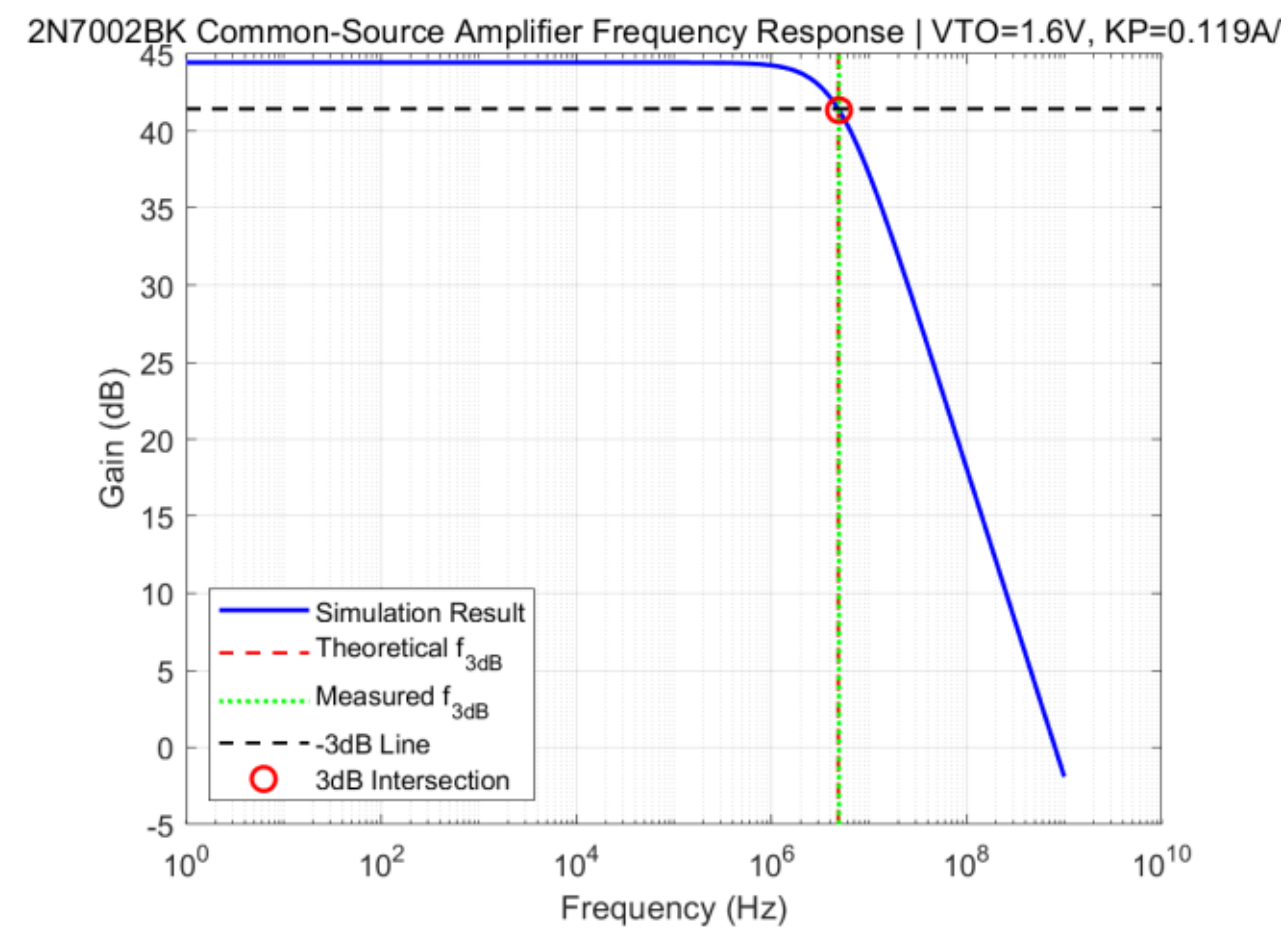

Fig. 13. Frequency response of the MOSFET common-source amplifier

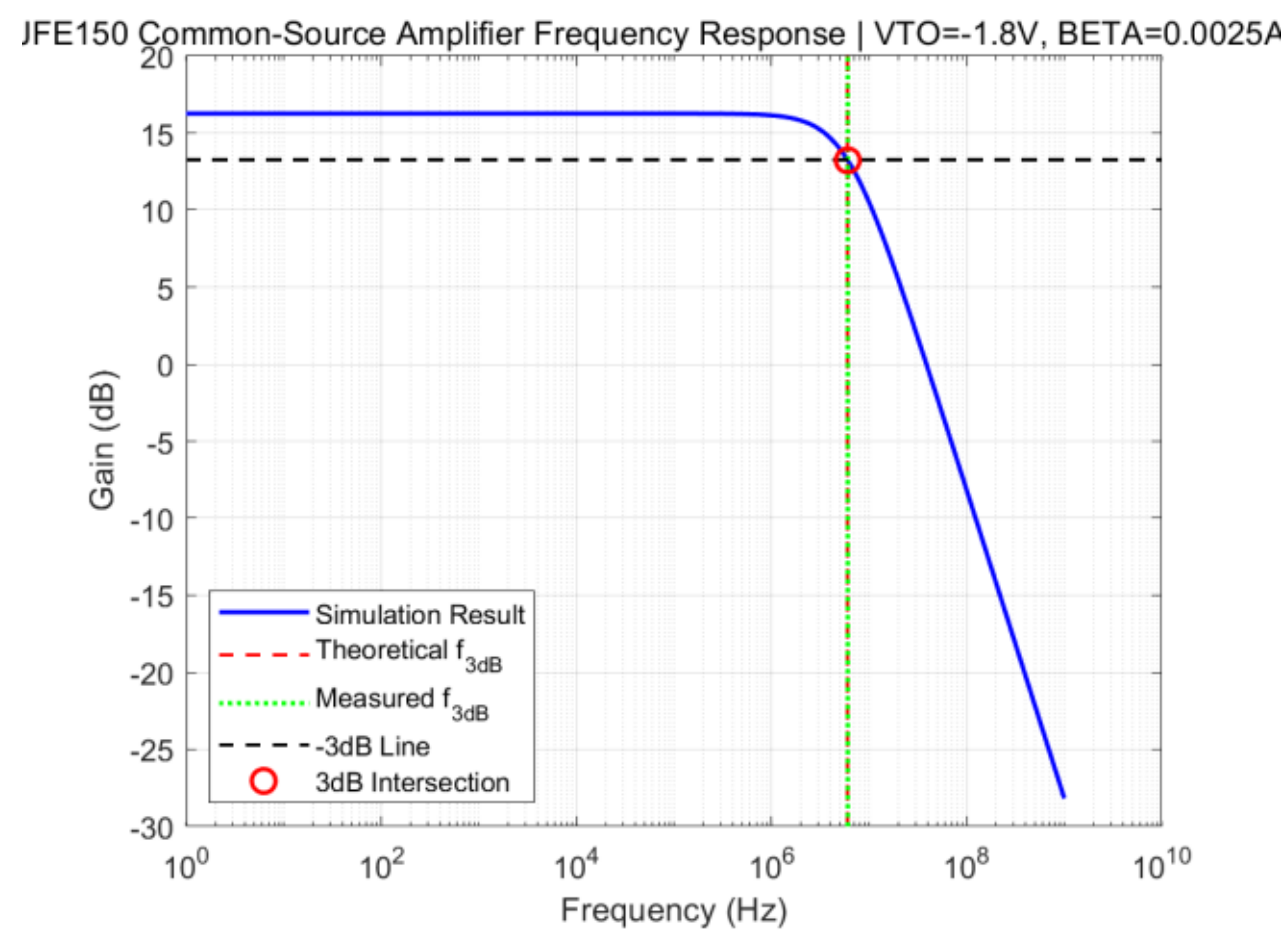

Fig. 14. Frequency response of the JFE150 JFET common-source amplifier

$$f_{theory} = (2\pi R C_j)^{-1} = (2\pi \times 1000 \times 1.02 \times 10^{-12})^{-1} = 155.56\text{MHz}$$

3) Simulation Setup: A MATLAB frequency response simulation (1 Hz to 1 GHz, 500 logarithmic points) measures the cutoff frequency fmeasured at the-3 dB point.

4) Calculate Measured Capacitance:

$$C_{j,measured} = (2\pi R f_{measured})^{-1}$$

$$C_{JO,measured} = C_{j,measured} \times \left(1 + \frac{V_R}{VJ}\right)^M$$

5) Error Analysis: Relative errors between theoretical and measured values are computed.

**Results:**

- Theoretical Cutoff Frequency: 155.56 MHz
- Measured Cutoff Frequency: 154.30 MHz
- Theoretical Cj: 1.02 pF
- Measured $C_{j,measured}$: 1.03 pF
- Input CJO: 4.00 pF
- Back-Calculated $C_{JO,measured}$: 4.03 pF

See Fig. 11. for frequency response curves.

**Conclusion:** Errors of 0.81% (cutoff frequency), 0.98% (Cj), and 0.75% (CJO) confirm the accuracy of CJO.

### E.2 Validation of BJT SPICE Model Parameters Using Frequency Response Analysis

**Objective:** This experiment validates the forward current gain BF = 100 and transit time TF = 640 ps in the SPICE model of BJTs (2N2218, 2N2219, 2N2221, 2N2222 series) from STMicroelectronics:

.model 2N2222 NPN(BF=100 TF=640e-12 CJC=8e-12 CJE=12e-12)

**Experimental Design:** Frequency-dependent current gain is analyzed, with theoretical low-frequency gain 20log10(BF) and transition frequency $f_T = (2\pi TF)^{-1}$.

**Experimental Procedure:**

1) Theoretical Calculations:

$$gain_{dB} = 20\log_{10}(current_{gain}) = 40.00\ dB$$

$$f_T = (2\pi \times 6.4 \times 10^{-10})^{-1} = 248.68\ MHz$$

2) Simulation Setup: MATLAB simulation (1 Hz to 1 GHz, 500 points) measures low-frequency gain and 0 dB crossover frequency f0dB.

3) Measure Transit Time:

$$TF_{measured} = (2\pi \times f_{0dB})^{-1}$$

4) Error Analysis: Relative errors are calculated.

**Results:**

- Theoretical Gain: 40.00 dB
- Measured Gain: 40.00 dB
- Theoretical fT: 248.68 MHz
- Measured $f_{0dB}$: 243.65 MHz

• Input TF: 640.00 ps
• Back-Calculated $TF_{measured}$: 653.20 ps
See Fig. 12. for frequency response.

**Conclusion:** Zero error in gain and 2.06% error in TF (within acceptable limits) validate the model parameters.

**E.3 Validation of MOSFET SPICE Model Parameters Using a Common Source Amplifier**

**Objective:** This experiment verifies V TO = 1.6 V, KP = 0.119 A/V2, CGSO = 28 pF, and CGDO = 5 pF in the SPICE model of the 2N7002BK MOSFET from Nexperia:

.model M1 NMOS(VTO=1.6 CGSO=28e-12 CGDO=5e-12 KP=0.119)

**Experimental Design:** A common-source amplifier (source grounded, 1 MΩ gate resistor, 1 kΩ drain resistor) measures low-frequency gain and 3 dB cutoff frequency.

**Experimental Procedure:**

1) DC Analysis: At VGS = 3V,

$$I_{D,theory} = K_P/2 \times (V_{GS} - V_{TO})^2 = 0.119/2 \times (3 - 1.6)^2 = 0.116\,A$$

2) Gain Calculation:

$$g_m = \sqrt{2 \times KP \times I_{D,theory}} = \sqrt{2 \times 0.119 \times 0.116 \times 10^{-3}} = 0.0166\,s$$

$$A_{v,theory} = -g_m \times R_D = -0.0166 \times 1000 = -16.6$$

3) Cutoff Frequency:

$$f_{3dB,theory} = (2\pi \times R_D \times (C_{GDO} + C_{GSO}))^{-1} = (2\pi \times 1000 \times (5 \times 10^{-12} + 28 \times 10^{-12}))^{-1} = 4.82\text{ MHz}$$

4) Simulation: MATLAB frequency response (1 Hz to 1 GHz, 500 points) measures gain and cutoff.

**Results:**
• Theoretical Gain: 24.40 dB
• Measured Gain: 24.38 dB
• Theoretical f3dB: 4.82 MHz
• Measured f3dB: 4.80 MHz
See Fig. 13. for frequency response.

**Conclusion:** Errors of 0.08% (gain) and 0.41% (f3dB) confirm the model parameters' accuracy.

**E.4 Validation of JFET SPICE Model Parameters Using a Common-Source Amplifier**

**Objective:** This experiment validates V TO = −1.8 V, BETA = 0.0025 A/V2, CGS = 24 pF, and CGD=2.5 pF in the SPICE model of the JFE150 JFET from Texas Instruments:

.model JFE150 NJF(VTO=-1.8 CGS=24e-12 CGD=2.5e-12 BETA=0.0025)

**Experimental Design:** A common-source amplifier (100 Ω source resistor, 1 MΩ gate resistor, 1 kΩ drain resistor) assesses gain and cutoff frequency.

**Experimental Procedure:**

1) DC Analysis: At VGS = −0.5V,

$$I_{D,theory} = \beta \times (V_{GS} - V_{TO})^2 = 0.0025 \times \left(-0.5 - (-1.8)\right)^2 = 0.00423\,A$$

2) Gain Calculation:

$$g_m = 2 \times \beta \times (V_{GS} - V_{TO}) = 2 \times 0.0025 \times 1.3 = 0.0065\,s$$

$$A_{v,theory} = -g_m \times R_D = -0.0065 \times 1000 = -6.5$$

3) Cutoff Frequency:

$$f_{3dB,theory} = \left(2\pi \times R_D \times (C_{GDO} + C_{GSO})\right)^{-1} = 6.01\text{ MHz}$$

4) Simulation: MATLAB frequency response (1 Hz to 1 GHz, 500 points) measures gain and cutoff.

**Results:**
• Theoretical Gain: 16.26 dB
• Measured Gain: 16.26 dB
• Theoretical f3dB: 6.01 MHz
• Measured f3dB: 6.05 MHz
See Fig. 14. for frequency response.

**Conclusion:** Zero error in gain and 0.67% error in f3dB validate the model parameters.

## APPENDIX F: WORKFLOW OF ABLATION STUDY VARIANTS

This appendix describes the full model's workflow and three ablation study variants, illustrated in Figs. 15 to 18.

- **Full Model Workflow (Fig. 15):** The full model's workflow integrates AGDF, HDER and HNEN.
- **Variant 1 (Fig. 16):** Removes attention-guided document focusing
  **Workflow**: Skips document selection, retrieves globally across the entire knowledge base.
  **Impact**: Increased noise, reduced accuracy due to irrelevant documents being included in the search.
- **Variant 2 (Fig. 17):** Removes hierarchical document-enhanced retrieval
  **Workflow**: No section prediction, performs full-document retrieval within the selected document.
  **Impact**: Lower precision due to the lack of focus on specific sections, leading to less targeted results.
- **Variant 3 (Fig. 18):** Removes heterogeneous named entity normalization
  **Workflow**: No standardization of parameter names, returns empty results if naming inconsistencies are encountered.
  **Impact**: Reduced robustness to naming inconsistencies, limiting the system's ability to handle diverse data sources.

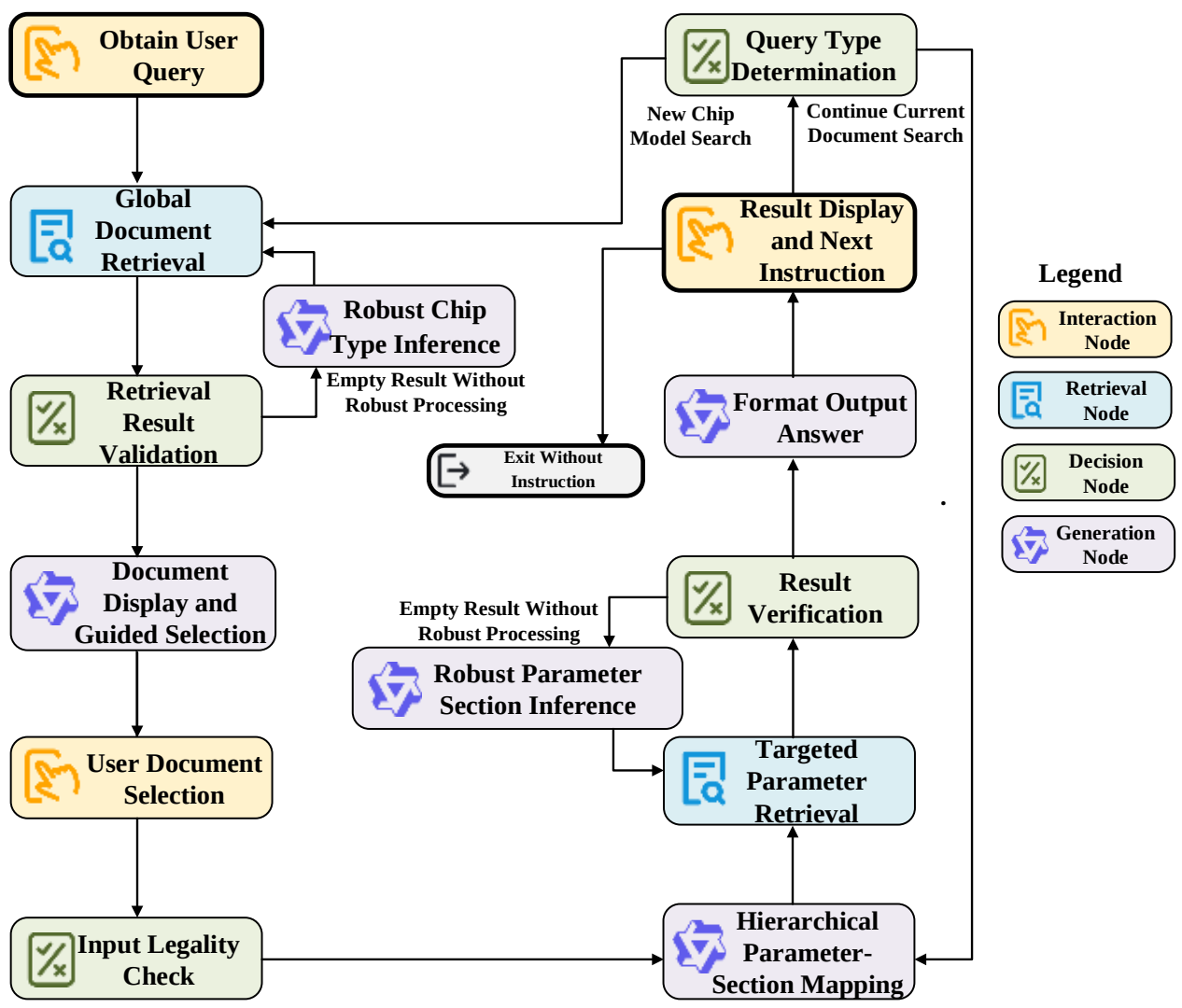

Fig. 15. Full Model Workflow

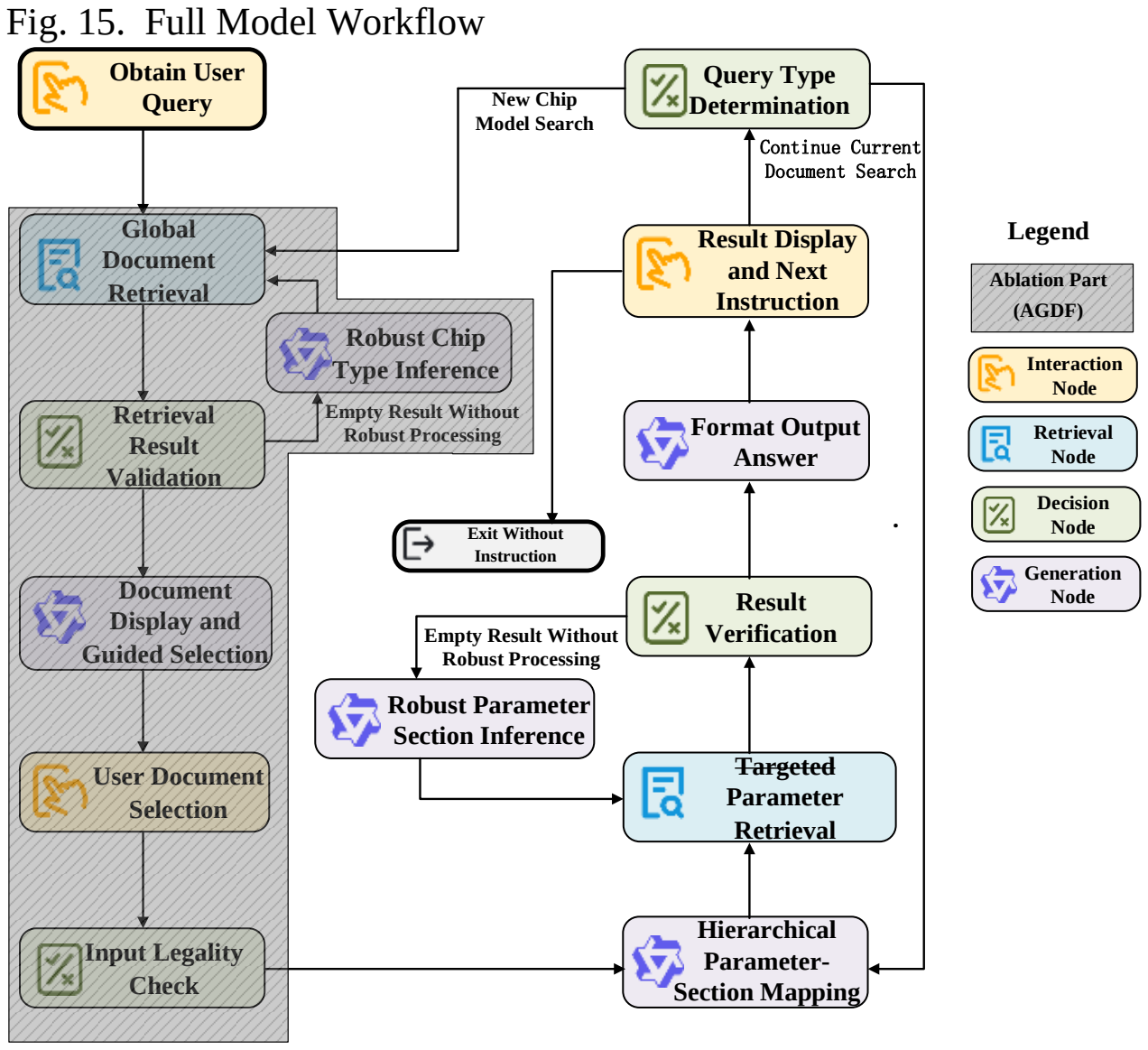

Fig. 16. Ablation Variant 1 - No Document Focusing.

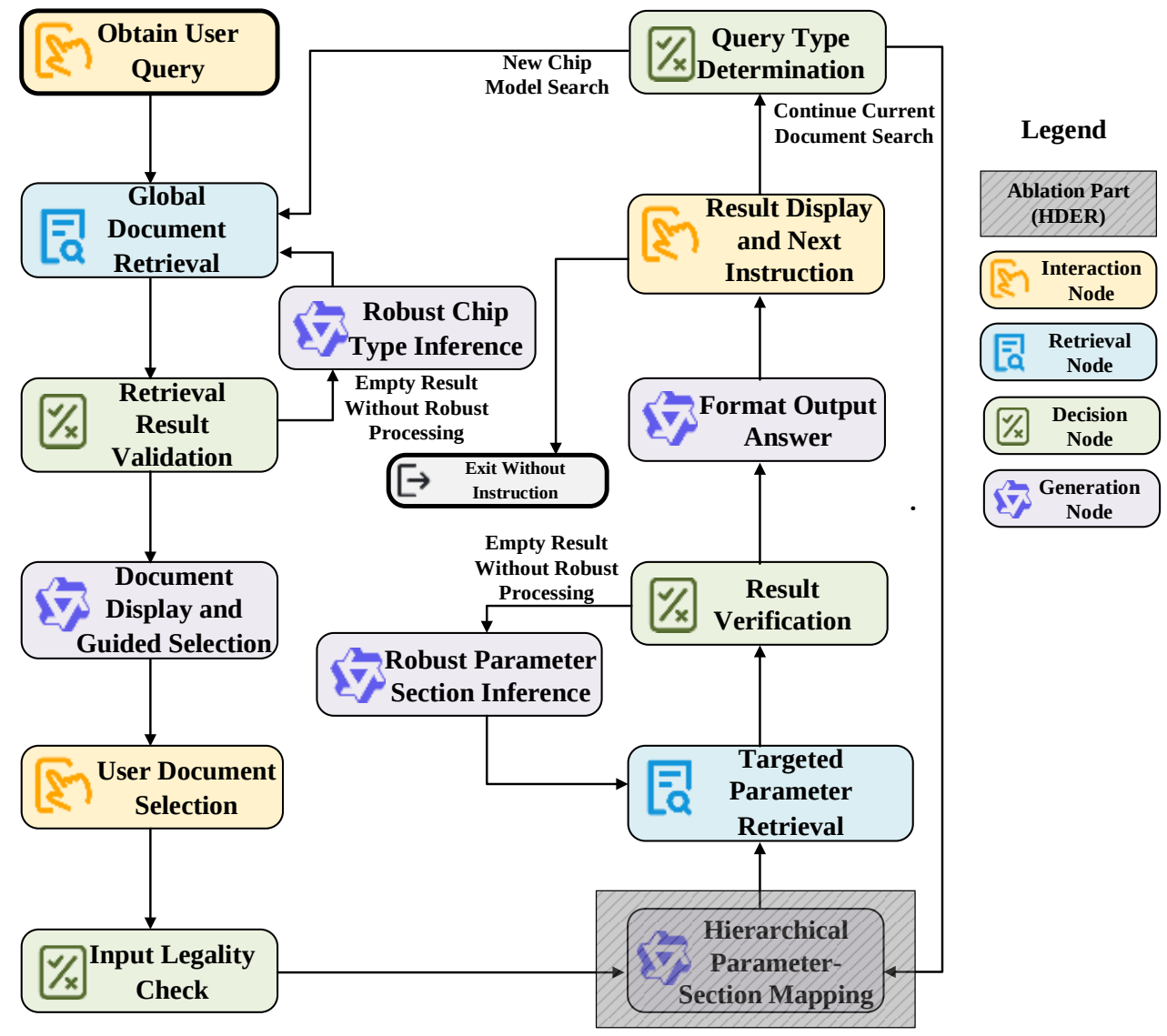

Fig. 17. Ablation Variant 2 - No Hierarchical Retrieval

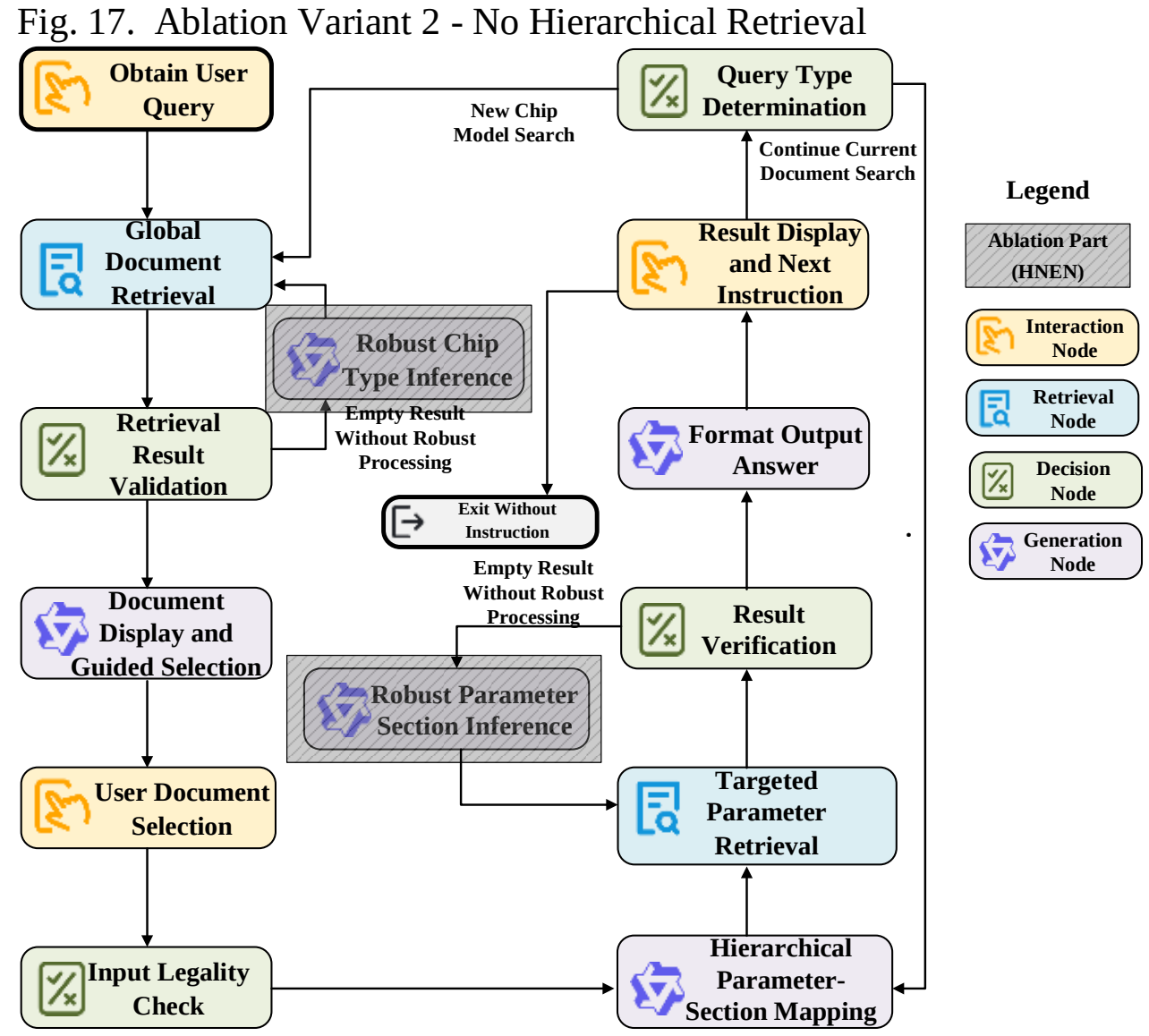

Fig. 18. Ablation Variant 3 - No Named Entity Normalization.